%% file: swe-opd.tex
\documentclass[dvipsnames]{article} %
\PassOptionsToPackage{table}{xcolor}
\usepackage{colm2024_conference}
\pdfoutput=1

\usepackage[utf8]{inputenc} % allow utf-8 input
\usepackage[T1]{fontenc}    % use 8-bit T1 fonts
\usepackage{xcolor}         % colors
\usepackage{url}
\usepackage{multicol}
\usepackage{multirow}
\usepackage{graphicx}
\usepackage{bm}
\usepackage{amsmath}
\usepackage{amssymb}
\usepackage{mathtools}
\usepackage{amsthm}
\usepackage{caption}
\usepackage{subcaption}
\usepackage{xspace}
\usepackage{bbding}
\usepackage{pifont}
\usepackage{colortbl}
\usepackage{enumitem}
\usepackage{overpic}
\usepackage{booktabs}
\usepackage{dsfont}
\usepackage{wrapfig}
\usepackage{needspace}
\usepackage{placeins}
\usepackage{float}
\usepackage[most]{tcolorbox}
\usepackage{fvextra}
\usepackage{tocloft}
\usepackage{algorithm}
\usepackage{algorithmic}

\input{math_commands.tex}

\definecolor{promptblue}{HTML}{24577A}
\definecolor{promptteal}{HTML}{28695D}
\newtcolorbox{agentprompt}[2]{enhanced,breakable,
  colback=#1!4!white,colframe=#1!75!black,colbacktitle=#1,
  coltitle=white,fonttitle=\small\bfseries,title={#2},
  title after break={#2 (continued)},
  boxrule=0.5pt,arc=1.5mm,left=8pt,right=8pt,top=6pt,bottom=6pt,
  before skip=8pt,after skip=10pt}
\makeatletter
\DeclareRobustCommand\onedot{\futurelet\@let@token\@onedot}
\def\@onedot{\ifx\@let@token.\else.\null\fi\xspace}

\newcommand\figcaption{\def\@captype{figure}\caption}
\newcommand\tabcaption{\def\@captype{table}\caption}
\makeatother

\renewcommand{\reportdate}{Nov 16, 2026}
\renewcommand{\runningtitle}{SWE-OPD: On-Policy Distillation for Agentic Coding}

\title{SWE-OPD: On-Policy Distillation for Agentic Coding}

\author{
\textbf{Yuhang He}$^{1, *}$,
\textbf{Yanyu Chen}$^{2, *}$,
\textbf{Chuanzhe Guo}$^{3, *}$,
\textbf{Jie Zhu}$^{4,\#,\dagger}$,
\textbf{Zichen Tian}$^{2}$,
\textbf{Hongyu Ge}$^{1}$,
\textbf{Zixuan Ren}$^{4}$,
\textbf{Huang He}$^{4}$,
\textbf{Bingjin Chen}$^{4}$,
\textbf{Junyuan Shang}$^{4}$,
\textbf{Shuohuan Wang}$^{4}$,
\textbf{Yongqi Zhang}$^{1,\dagger}$,
\textbf{Irwin King}$^{2,\dagger}$,
\textbf{Jing Liu}$^{4}$ \\
\vspace{0.1cm}
{\small
$^{1}$The Hong Kong University of Science and Technology (Guangzhou) \\
$^{2}$The Chinese University of Hong Kong \quad
$^{3}$Harbin Institute of Technology \quad \\
$^{4}$ERNIE Team, Baidu \\
 ${^*}$ Equal Contribution \quad 
$^{\#}$Project Lead \quad
$^{\dagger}$Corresponding Author \\
}
\quad \\
{\footnotesize
Email: zhujie13@baidu.com
}
}

\makeatletter\def\@abstract{
In this work, we study on-policy distillation (OPD) for transferring agentic coding capabilities from a large teacher model to a \textit{smaller} student model. Prior work has focused primarily on reasoning and short-horizon agentic tasks. We present SWE-OPD, a minimal OPD study for software engineering, where the student interacts with executable repositories and the teacher supervises the prefixes visited by the student. Across three teacher--student pairs, two training sources, and three SWE benchmarks, we find that on-policy supervision transfers coding capability: supervision from Qwen3.6-27B increases the SWE-bench Pro resolve rate of Qwen3.5-9B from 35.9\% to 39.4\%. Refreshing trajectories is not strictly required either, as an offline variant with cached teacher scores still improves over the student baselines while cutting end-to-end GPU-hours by 59.4\%--72.6\% and halving peak GPU allocation, at benchmark-dependent accuracy. Furthermore, we extend the offline variant to corss-tokenizer scenarios: distilling GLM-4.5-Air from DeepSeek-V4-Flash-0731 raises its SWE-bench Pro and Multilingual resolve rate from 35.8\% to 39.7\% and from 41.8\% to 46.2\%, around 4\% and 4.6\% improvement, respectively. We also find that a decreasing loss, however, does not necessarily indicate improved agent performance: under top-$K$ reverse-KL loss, one student exhibits severe repetition despite a decreasing loss. We identify a support-mass gradient term as a potential mechanism underlying this failure.\\

\textbf{Code: \url{https://github.com/ernie-research/SWE-OPD}}
}\makeatother

\begin{document}

\maketitle

\newpage

\clearpage
\pagestyle{firstpage}  %
\tableofcontents
\clearpage
\pagestyle{normalpage}

\section{Introduction}

On-policy distillation (OPD)~\citep{li2026rethinking, agarwal2024policy, lu2025onpolicydistillation, yang2026learning, zhao2026self, yang2026nemotron, song2026survey, wu2026lightning, gu2023minillm} transfers the capabilities of a strong teacher to a student through token-level supervision on trajectories generated by the student. 
This distinguishes OPD from conventional knowledge distillation and supervised fine-tuning~\citep{gou2021knowledge, hinton2015distilling, ye2025analyzing}, which typically rely on fixed, teacher-generated responses. 
Accordingly, OPD has recently attracted growing attention, with applications ranging from reasoning~\citep{li2026rethinking, xing2026trust, yang2026learning, yang2026self, hou2026uni, wu2026lightning} to relatively simple agentic tasks such as web search~\citep{zhou2026sage, wang2026tcod, wang2026mad}~\footnote{In industry, OPD is viewed as an effective capability-merging method~\citep{team2026kimi, yang2025qwen3, xiao2026mimo, ma2026mopd, zeng2026glm}, enabling the integration of specialized capabilities from multiple teachers (usually derived from reinforcement learning) into a single student of the \textit{same} size.}. 
Nevertheless, the effectiveness of OPD in long-horizon tasks that require interaction with an environment, such as software engineering (SWE), remains underexplored.

However, dense supervision alone does not guarantee that OPD will improve SWE agents. An erroneous action can modify repository contents or alter the observations that inform subsequent decisions, requiring the teacher to score histories that differ from those generated under its own policy.  
Moreover, long interaction histories make teacher scoring costly. Practical approximations, such as restricting scored candidates or training on fixed trajectories, change how supervision is used or where it is applied. These considerations make SWE a demanding setting for examining both the effectiveness and the limitations of on-policy teacher supervision. Recent analyses of OPD also highlight the importance of reliable supervision over long trajectories~\citep{li2026rethinking}.

We investigate these questions through SWE-OPD, a minimal distillation setup for coding agents. A \textit{small} student interacts with executable repositories, while a \textit{larger} teacher scores the student-visited prefixes.  
We compare a top-$K$ reverse-KL loss with a sampled-token objective to examine whether the choice of candidate-level supervision affects task performance and training behavior. 
For simplicity, we use vanilla OPD and retain valid unresolved and budget-truncated trajectories, allowing us to evaluate distillation without requiring successful student demonstrations for every training example.

We further examine whether effective distillation requires continually refreshed student trajectories. Following the offline approach explored in Lightning OPD~\citep{wu2026lightning}, Offline SWE-OPD collects trajectories from the initial student and precomputes teacher supervision before optimization. This decouples rollout collection, teacher scoring, and student training, allowing the stages to share a GPU pool. The fixed offline trajectory distribution may diverge from the distribution induced by the evolving student policy. Hence, it is necessary to evaluate the offline variant through both downstream resolve rates and end-to-end resource use.

Across three teacher-student pairs, two training sources, and three SWE benchmarks, our experiments demonstrate that vanilla OPD can improve instruction-tuned coding agents. 
For example, distillation on ScaleSWE with Qwen3.6-27B as the teacher increases the SWE-bench Pro resolve rate of Qwen3.5-9B from 35.9\% to 39.4\%.
In our internal-data experiments, offline SWE-OPD reduces end-to-end GPU-hours by 59.4\% for Qwen and 72.6\% for GLM while also bringing improvement. The offline pipeline also accepts a teacher with a different tokenizer: merging generated tokens whose spans differ between the two tokenizations following the simplest method GOLD~\citep{2025_unlocking_on_policy_distillation_for_any_model_family}, GLM-4.5-Air distilled from DeepSeek-V4-Flash-0731 improves on all three benchmarks.
These results indicate a favorable trade-off between task performance and computational cost under the evaluated training budgets.
Our experiments also reveal a limitation of evaluating distillation solely through its training objective. Although the top-K loss improves the Qwen student, the GLM student develops severe repetition even as its training loss decreases. In contrast, sampled-token training remains stable in the evaluated configuration. An analysis of the implemented top-K loss identifies a support-mass gradient that can shift probability outside the scored candidates, even at teacher-student agreement. This provides a potential failure mechanism for the top-K loss. Together, these findings motivate evaluating SWE distillation through task success, training behavior, and resource cost, and distinguishing supervision coverage from the reliability of the resulting updates. Our contributions are summarized:
\begin{itemize}[leftmargin=*, nosep]
\item  \textbf{Capability transfer in agentic coding.} Our controlled study of long-horizon coding shows that teacher supervision on student-visited prefixes can transfer capabilities across model scales and tokenizers, improving coding agents without reinforcement learning or filtering trajectories by task success.  

\item  \textbf{Learning without trajectory refresh.} We show that online trajectory refresh is not required for improvement. Using cached teacher supervision on fixed trajectories collected from the initial student halves GPU-hours while maintaining satisfactory performance relative to online training.    

\item  \textbf{Reliability beyond loss reduction.} We demonstrate that optimizing the top-K loss can worsen alignment with the teacher even as the loss decreases. It exposes a structural risk of shifting probability outside scored candidates, inspiring furture works on reconsidering OPD objectives.

\end{itemize}
\textbf{Scope of this study.} This work is designed as a controlled study of vanilla OPD for SWE. We intentionally avoid combining OPD with reinforcement learning, data filtering, or specialized optimization techniques, allowing us to examine whether on-policy teacher supervision alone can transfer agentic coding capabilities across model scales. Therefore we believe that there remains room for further improving student performance, and we release our code to facilitate future exploration.

% The remainder of this paper is organized as follows. Section 3 defines the task and the study design; Section 4 presents on-policy distillation for SWE and its capability-transfer results; Section 5 studies offline supervision, its performance and its cost; Section 6 examines supervision coverage and training stability.

\section{Related Work}

Classical supervision learning is trained
only on teacher-visited prefixes~\citep{kim2016sequence}, while at test time the student must condition on its own,
so errors compound along the trajectory. On-policy distillation (OPD) closes this gap by sampling
trajectories from the student and letting the teacher score them, inheriting the
on-/off-policy distinction from policy distillation for RL agents~\citep{rusu2016policy}.
\citet{lin2020autoregressive} first brought DAgger-style~\citep{DBLP:journals/jmlr/RossGB11} rollouts to autoregressive
distillation; MiniLLM~\citep{gu2023minillm} minimises sequence-level reverse KL and shows it
is equivalent to a policy gradient whose reward is the teacher--student log-ratio; GKD
\citep{agarwal2024gkd} unifies the resulting design space over data mixtures and divergences. 

Compared with reinforcement learning from verifiable rewards, where an entire rollout
collapses into a single scalar~\citep{shao2024deepseekmath,deepseekr1}, the appeal of OPD is
the density of its signal: every sampled token receives gradient, which buys substantial
compute savings at comparable accuracy~\citep{lu2025onpolicydistillation}. Existing methods vary along
three axes~\citep{song2026survey}: the divergence (JSD in GKD~\citep{DBLP:journals/corr/abs-2306-13649}, skewed KL in DistiLLM
\citep{ko2024distillm}, reverse KL where mode-seeking behaviour helps on reasoning tasks);
the granularity of teacher access (the log-probability of the single sampled token, the
teacher's top-$K$, or the full vocabulary); and whether the teacher is an external model or
the student itself under privileged context~\citep{zhao2026self}. 

However, the reported gains are nonetheless concentrated on reasoning and relatively simple agentic tasks, such as competitive mathematics. In this work, we extend OPD to more complex agentic coding scenarios, where solving tasks requires long-horizon planning, iterative tool use, and sustained interaction with execution environments. We investigate whether OPD can effectively transfer capabilities from stronger teacher models to smaller coding agents under these more challenging settings.

\section{Preliminaries}
\label{sec:setup}

%\subsection{Agentic Coding}

We consider SWE tasks as sequential decision-making problems in executable software environments.
A task is specified by $x=(\mathcal{I},\mathcal{R})$, comprising an issue $\mathcal{I}$ and a repository $\mathcal{R}$. At turn $t$, $s_t$ denotes the full state of environment $\mathcal{E}_{\mathcal{R}}$, including the repository contents and execution context, which is only partially observable. The goal is to produce a patch to fix the issue.

The agent policy $\pi_\theta$ generates an assistant response $a_t$ conditioned on the task prompt and the interaction history.
Each response may contain reasoning, tool invocations, or a final answer.
The environment executes the tool invocations in $a_t$ against the current state $s_t$, returns an observation $o_t=\mathcal{E}_{\mathcal{R}}(s_t,a_t)$ such as file contents or test logs, and transitions to $s_{t+1}$.
The interaction terminates when the agent submits its result or exhausts a predefined interaction budget, yielding the trajectory $\tau=(x,a_0,o_0,a_1,o_1,\ldots,a_T,o_T)$.
We allow $o_t$ to be empty when $a_t$ contains no tool invocation. 
After termination, the resulting patch is applied to the initial repository and evaluated using the benchmark's required regression tests.
A task is considered \emph{resolved} only if all required tests pass.
In on-policy distillation, a student generates trajectories and a frozen teacher provides next-token probabilities conditioned on the prefixes visited by the student. The student is then updated using discrepancies between the two models' predictions on these prefixes.
Throughout this work, we sample only one single trajectory per training prompt $x$.

\begin{figure}[t]
\centering
\includegraphics[width=0.95\linewidth]{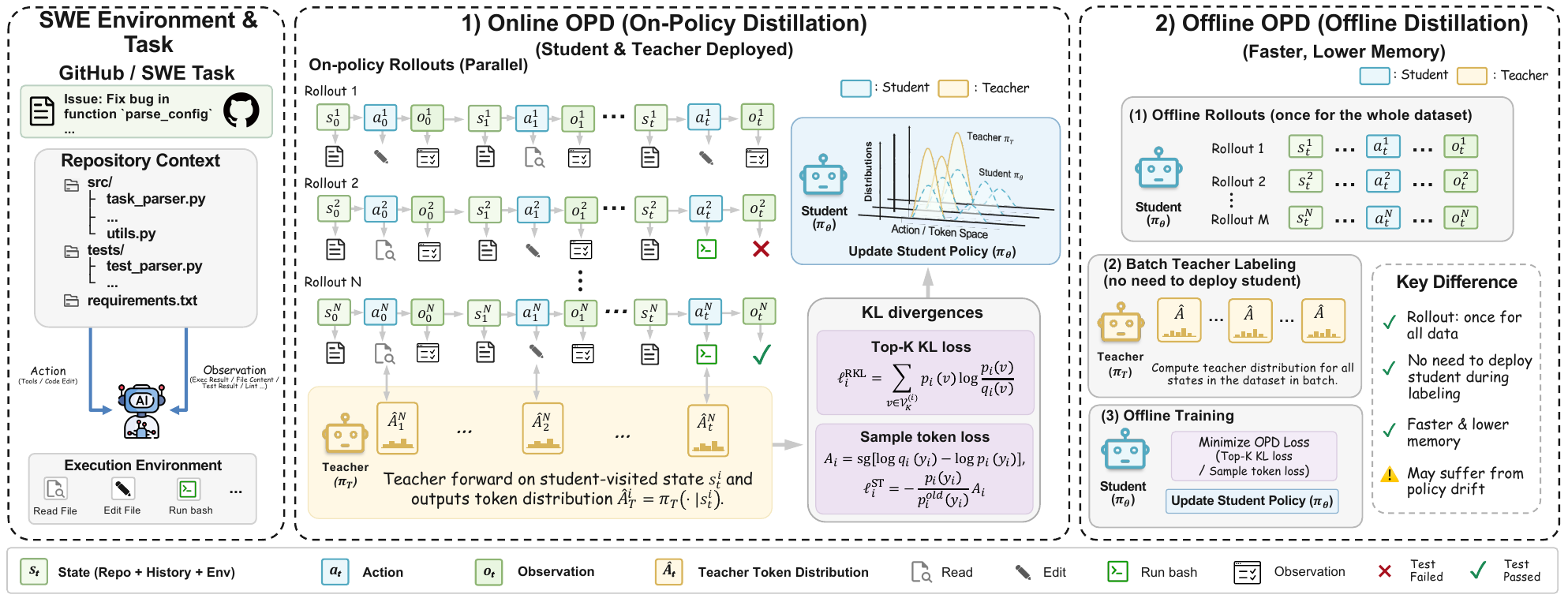}
\caption{\textbf{Study design of SWE-OPD}. \textbf{Left:} the student interacts with an executable repository; tool observations enter the conditioning context, while only student-generated tokens receive supervision. \textbf{Middle \& Right:} The two design axes we compare are the \emph{trajectory source} (current student in SWE-OPD vs.\ initial student in Offline SWE-OPD) and the \emph{teacher information} used per position (top-$K$ candidates in the RKL loss vs.\ the sampled token in the ST loss objective).
}
\label{fig:main-pipeline}
\end{figure}
% main fig 后续继续更新
\section{SWE-OPD: On-Policy Distillation for SWE}
\label{sec:vanilla}

SWE-OPD follows standard multi-turn OPD, conditioning teacher supervision on the student's decisions and observations in executable repositories (Figure~\ref{fig:main-pipeline} Left). We define rollout collection (Section~\ref{sec:rollout}), frozen-teacher scoring (Section~\ref{sec:teacher}), and two objectives with different amounts of candidate-level teacher information (Section~\ref{sec:objective}), then evaluate end-to-end issue resoution as shown in Figure~\ref{fig:main-pipeline} Middle.

% Even though, we deliberately adopt the standard multi-turn OPD recipe, rather than introduce a new OPD estimator~\citep{zhou2026turnopd, team2026agents}.  Our SWE-OPD comprises three components. The \textbf{rollout} component executes the student policy $\pi_\theta$ in a repository (Section~\ref{sec:rollout}). The frozen \textbf{teacher} policy $\pi_T$ scores student-generated positions without producing a separate trajectory (Section~\ref{sec:teacher}). The \textbf{trainer} then updates $\theta$ using the resulting token-level discrepancies (Section~\ref{sec:objective}).
\subsection{Rollout in Executable Repositories}
\label{sec:rollout}

For each task $x$, the current student $\pi_\theta$ generates a repository trajectory following Section~\ref{sec:setup}. Its actions determine the visited states and observations in subsequent prefixes used for teacher scoring. To study vanilla OPD without additional optimization strategies, we retain all valid rollouts, including unresolved and budget-truncated episodes, and leave selective data curation to future work.

\subsection{Teacher Scoring on Student-Visited Prefixes}
\label{sec:teacher}

The frozen teacher provides next-token probabilities on the prefixes visited by the student. Importantly, The teacher and student condition on the same observable prefix, comprising the issue description, previous responses, and tool outputs.
For token-level supervision, we represent the responses and observations in $\tau$ following $x$ as a single token sequence $\mathbf{y}=(y_1,\ldots,y_L)$ in interaction order. Let $\mathcal{G}(\tau)$ denote all positions corresponding to student-generated response tokens, and define the generation mask as $m_i=\mathds{1}[i\in\mathcal{G}(\tau)]$.
At a generated position $i$ within response $a_t$, the prefix $(x,\mathbf{y}_{<i})$ contains the task context, preceding responses and observations, and earlier tokens in the current response. Unless stated otherwise, the student and teacher share a tokenizer and token-ID mapping, with vocabulary $\mathcal V$; Section~\ref{sec:cross-tokenizer} relaxes this assumption for offline ST training.
We define their next-token probabilities on the shared prefix:
\begin{equation}
p_i(v) = \pi_\theta(v\mid x,\mathbf{y}_{<i}),
\qquad
q_i(v) = \pi_T(v\mid x,\mathbf{y}_{<i}).
\label{eq:shared-prefix}
\end{equation}
We apply token-level supervision at every student-generated position $i \in \mathcal{G}(\tau)$. Tool outputs, test logs, and retrieved repository contents remain part of the conditioning context but are excluded from the loss by the generation mask $m_i$. Supervision is therefore dense across generated positions.
For training with the top-$K$ reverse KL loss, we retain teacher log-probabilities for the current student's $K$ highest-probability candidates, rather than teacher scores over the full vocabulary:
\begin{equation}
\mathcal{V}_K^{(i)}=\operatorname{TopK}\!\left(p_i,K\right) \, .
\label{eq:student-topk}
\end{equation}

Retaining $K$ teacher scores per generated position gives $O(|\mathcal{G}(\tau)|K)$ scores per trajectory, reducing storage and communication for coding histories that can exceed $128k$ tokens. The ST loss (Eq.~\ref{eq:sampled-token-loss}) instead uses the teacher score of the generated token $y_i$, even outside $\mathcal{V}_K$. Both variants supervise the same positions with different candidate-level information (Section~\ref{sec:objective}).
\subsection{Student Optimization via Distillation}
\label{sec:objective}

Our goal is to align the student's next-token predictions with the teacher on student-visited prefixes. We consider the top-$K$ reverse KL loss (RKL) and the sampled-token policy-gradient loss (ST). 
For a batch of $B$ trajectories $\tau_1,\ldots,\tau_B$, the set $\mathcal{G}_B=\{(b,i):1\leq b\leq B,\ i\in\mathcal{G}(\tau_b)\}$ collects all supervised positions.
We average the loss over these positions:
\begin{equation}
\mathcal{L}_{\mathrm{OPD}}
=\frac{1}{|\mathcal{G}_B|}
\sum_{(b,i)\in\mathcal{G}_B}\ell_{b,i}^{*},
\qquad *\in\{\mathrm{RKL},\mathrm{ST}\}.
\label{eq:swe-opd-objective}
\end{equation}
We instantiate $\ell_{b,i}^{*}$ with either the RKL or ST loss below, omitting the batch index $b$ for clarity.
\paragraph{Top-$K$ reverse-KL loss.}
On the current student's top-$K$ support $\mathcal{V}_K^{(i)}$, we follow the protocol of~\citet{li2026rethinking} and normalize student weights as $\bar p_i(v)=p_i(v)/\Zeta_i^p$ for $v\in\mathcal{V}_K^{(i)}$, where $\Zeta_i^p=\sum_{v\in\mathcal{V}_K^{(i)}}\operatorname{sg}[p_i(v)]$ ($\operatorname{sg}$ denotes the stop-gradient operation). The token-level loss is
\begin{equation}
\ell_i^{\mathrm{RKL}}
=
\sum_{v\in\mathcal{V}_K^{(i)}}
\bar p_i(v)\log\frac{p_i(v)}{q_i(v)}.
\label{eq:rkl-topk}
\end{equation}
This is a surrogate for full-vocabulary reverse KL: the log-ratio uses the original probabilities, while the student weights are normalized over the selected support.
The support and its normalization constant are recomputed at each update and held fixed during differentiation; gradients flow through both $\bar p_i(v)$ and $\log p_i(v)$.
The objective is motivated by the \emph{mode-seeking} intuition of reverse KL: encouraging the student to concentrate probability on continuations favored by the teacher~\citep{gu2023minillm}.
A candidate contributes a large positive loss when the student assigns it substantial probability but the teacher assigns it much lower probability.
Restricting the sum to the student's top-$K$ candidates focuses this penalty on continuations the student is most likely to generate.
This motivates its use in long-horizon coding, where an erroneous early action can alter the repository state and affect subsequent decisions.
\footnote{We also consider the forward-KL loss $\ell_i^{\mathrm{FKL}}=\mathbb{E}_{v\sim\bar q_i}[\log(q_i(v)/p_i(v))]$ in our preliminary experiments. Its performance is inferior to that of the RKL loss, so we mainly focus on the latter.}

\paragraph{Sampled-token loss.}
The ST loss uses the teacher--student log-probability ratio at the sampled token as a token-level advantage, following~\cite{ma2026mopd} and~\cite{li2026rethinking}.
Let $p_i^{\mathrm{old}}(v)=\pi_{\theta_{\mathrm{old}}}(v\mid x,\mathbf{y}_{<i})$ denote the frozen student policy used to collect the rollout.
For $y_i\sim p_i^{\mathrm{old}}$, we define
\begin{equation}
A_i =
\operatorname{sg}\!\left[
\log q_i(y_i)-\log p_i(y_i)
\right],
\qquad
\ell_i^{\mathrm{ST}}
=
-\frac{p_i(y_i)}{p_i^{\mathrm{old}}(y_i)}A_i.
\label{eq:sampled-token-loss}
\end{equation}
The advantage is recomputed using the current student and held fixed during differentiation. 
It encourages increasing $p_i(y_i)$ when $q_i(y_i)>p_i(y_i)$, and decreasing it when the inequality is reversed.
The token is sampled from the rollout policy; this objective is therefore distinct from setting $K=1$ in Eq.~\ref{eq:rkl-topk}. 
It requires only the teacher log-probability $\log q_i(y_i)$ at each generated position and estimates the gradient using a single Monte Carlo sample.
%and the unbiasedness statement in Appendix~\ref{app:wf-sampled-token} applies only to the unclipped, fixed-prefix form.

\subsection{Capability Transfer to Coding Agents}
\label{sec:opd-results}

\begin{table}[htbp]
\centering
\caption[Results on ScaleSWE.]{Results after training on 5K ScaleSWE queries. Avg@4 is the mean resolve rate (\%) over four independent runs on SWE-bench Verified~\citep{jimenez2024swe}, Pro~\citep{deng2025swe}, and Multilingual~\citep{zan2026multi}. $\Delta$ is the change from the student baseline in percentage points; bold marks the best score per benchmark. Subsequent benchmark tables use these conventions within each student--teacher pair.\protect\footnotemark}
\label{tab:main}
\small
\begin{tabular*}{\linewidth}{@{\extracolsep{\fill}}lcccccc}
\toprule
\multirow{2}{*}{Method}
& \multicolumn{2}{c}{Verified}
& \multicolumn{2}{c}{Pro}
& \multicolumn{2}{c}{Multilingual} \\
\cmidrule(lr){2-3}\cmidrule(lr){4-5}\cmidrule(lr){6-7}
& Score & $\Delta$ & Score & $\Delta$ & Score & $\Delta$ \\
\midrule
\multicolumn{7}{l}{\emph{Student: Qwen3.5-9B \quad Teacher: Qwen3.6-27B}} \\
\quad Student baseline & 57.9 &  & 35.9 &  & 43.0 &  \\
\quad Teacher Rollout Distillation & 54.8 & $-3.1$ & 34.5 & $-1.4$ & 40.5 & $-2.5$ \\
\quad RL (UniAgent~\citep{uniagent_github}) & 59.2 & $+1.3$ & -- &  & -- &  \\
\quad SWE-OPD (RKL) & {60.5} & $+2.6$ & \textbf{39.4} & $+3.5$ & \textbf{45.8} & $+2.8$ \\
\quad SWE-OPD (ST) & \textbf{61.7} & $+3.8$ & 37.9 & $+2.0$ & 45.2 & $+2.2$ \\
\bottomrule
\end{tabular*}

% \vspace{0.4em}
% {\footnotesize \emph{Setting.} $T{=}1.0$, $\text{top-}p{=}0.95$, 128K context, 100 max step, and one attempt per instance.}
\end{table}
\footnotetext{The experimental setup, training configurations, and agent scaffold are detailed in Appendix~\ref{app:implementation}.}

\textbf{Does teacher supervision on student-visited prefixes transfer coding capability?} Table~\ref{tab:main} reports results for Qwen3.5-9B trained with supervision from Qwen3.6-27B on 5K ScaleSWE training queries~\citep{zhao2026immersiongithubuniversescaling}~\footnote{We use instruct models as companies increasingly release fewer base models, though the method also applies to base models. Results may differ from official reports because detailed evaluation settings, including harnesses and scaffolds, are unavailable.}.
We hold the training queries, agent scaffold, and rollout budget fixed while comparing three training configurations: SFT on teacher rollouts (Teacher Rollout Distillation), OPD with the RKL objective, and OPD with the ST objective.
Teacher Rollout Distillation retains unresolved and truncated teacher episodes without reward filtering, matching the trajectory-retention policy of SWE-OPD  (Section~\ref{sec:rollout}).
Both OPD objectives improve the student baseline on all three benchmarks: RKL gains 2.6\%, 3.5\%, and 2.8\% on Verified, Pro, and Multilingual, respectively, while ST gains 3.8\%, 2.0\%, and 2.2\%. Teacher Rollout Distillation does not improve the student and degrades it on all three benchmarks ($-3.1$\%, $-1.4$\%, and $-2.5$\%). 
Under this shared trajectory-retention policy, the comparison examines differences in both the source and form of supervision: teacher-generated tokens on teacher-visited prefixes versus teacher scores on student-visited prefixes---rather than the effect of data curation. 
The degradation is consistent with distribution shift: fine-tuning on teacher-visited prefixes may distort an instruction-tuned student's distribution more than dense supervision on its own prefixes. Rejection-sampling SFT on resolved teacher trajectories would provide a stronger off-policy baseline and likely recover part of the gap; we leave this comparison to future work. Our results therefore support OPD over off-policy distillation under matched, unfiltered data, without establishing superiority over the available SFT pipeline.

% Additionally, OPD exhibits strong cross-benchmark generalization despite the narrow domain of its training data. Our training set is constructed from ScaleSWE and contains only Python-related tasks, yet the resulting model improves not only on SWE-bench Verified, but also on SWE-bench Pro and SWE-bench Multilingual. This suggests that OPD can transfer coding capabilities learned from a relatively narrow training distribution to substantially different task distributions.

\textbf{Do the gains persist with another training source and a larger student?} We repeat the comparison on approximately
4K internal training tasks~\citep{guo2026menvagent} and report both student--teacher pairs in Table~\ref{tab:internal}. 
For Qwen3.5-9B, RKL again improves all three benchmarks,
while ST improves Verified and Pro with little change
on Multilingual.
For the 106B-parameter GLM-4.5-Air student, ST improves
all three benchmarks, especially with a 3.4\% gain
on Multilingual.
Within the studied budgets, these results indicate that OPD can further
improve a large instruction-tuned coding agent, and that the choice of
objective matters: RKL training collapses for GLM despite its gains for
Qwen, which we examine in Section~\ref{sec:stability}.

\begin{table}[htbp]
\small
\centering
\caption{Results on the internal dataset. We evaluate SWE-OPD with Qwen3.5-9B and GLM-4.5-Air as students, supervised by Qwen3.6-27B and GLM-5.2, respectively.}
\label{tab:internal}
\begin{tabular*}{\linewidth}{@{\extracolsep{\fill}}lcccccc}
\toprule
\multirow{2}{*}{Method}
& \multicolumn{2}{c}{Verified}
& \multicolumn{2}{c}{Pro}
& \multicolumn{2}{c}{Multilingual} \\
\cmidrule(lr){2-3}\cmidrule(lr){4-5}\cmidrule(lr){6-7}
& Score & $\Delta$ & Score & $\Delta$ & Score & $\Delta$ \\
\midrule
\multicolumn{7}{l}{\emph{Student: Qwen3.5-9B \quad Teacher: Qwen3.6-27B}} \\
\quad Student baseline & 57.9 &  & 35.9 &  & 43.0 &  \\
\quad SWE-OPD (RKL) & \textbf{61.6} & $+3.7$ & 38.4 & $+2.5$ & \textbf{44.3} & $+1.3$ \\
% ST, internal data, four evaluation runs:
% Verified: 60.4 / 62.2 / 62.6 / 60.0; mean = 61.3.
% Pro: 39.1 / 38.4 / 39.4 / 37.8; mean = 38.675.
% Multilingual: 43.3 / 43.3 / 42.9 / 42.7; mean = 43.05.
% Displayed means use round-half-up to one decimal place.
% Gains use the unrounded benchmark means: RKL = 2.3 / 2.7 / 1.2; ST = 3.4 / 2.775 / 0.05.
\quad SWE-OPD (ST) & {61.3} & $+3.4$ & \textbf{38.7} & $+2.8$ & 43.1 & $+0.1$ \\
\midrule
\multicolumn{7}{l}{\emph{Student: GLM-4.5-Air \quad Teacher: GLM-5.2}} \\
\quad Student baseline & 59.9 &  & 35.8 &  & 41.6 &  \\
% GLM RKL training collapses; see Section~\ref{sec:why-topk-fails}.
% \quad SWE-OPD (RKL)$^\dagger$ & N/A &  & N/A &  & N/A &  \\
\quad SWE-OPD (ST)$^\dagger$ & \textbf{60.4} & $+0.5$ & \textbf{37.1} & $+1.3$ & \textbf{45.0} & $+3.4$ \\
% & \textbf{60.4} & 37.1 & \textbf{45.0}
\bottomrule
\end{tabular*}
\par\vspace{3pt}
\begin{minipage}{\linewidth}
\footnotesize\raggedright
$^\dagger$ RKL training on GLM collapses with severe repetition; see Section~\ref{sec:why-topk-fails}.
\end{minipage}
\end{table}

\subsection{Top-K Selection and Support Analysis}
\label{sec:coverage}
\label{sec:topk-analysis}

\begingroup
\captionsetup{subrefformat=simple}
\begin{wrapfigure}{r}{0.30\linewidth}
\centering
\includegraphics[width=\linewidth]{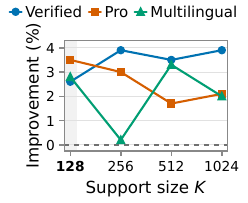}
\captionsetup{font=footnotesize,justification=raggedright,singlelinecheck=false}
\caption{\textbf{Top-$K$ ablation.} Gains in percentage points.}
\label{fig:topk-ablation}
\end{wrapfigure}

\textbf{How large must the top-K be?} We sweep $K\in\{128,256,512,1024\}$ for one epoch on ScaleSWE, distilling Qwen3.5-9B from Qwen3.6-27B with the RKL loss.
Gains over the student baseline persist across all tested support sizes and all three benchmarks, with no consistent benefit from increasing $K$ (Figure~\ref{fig:topk-ablation}).
% All settings improve on the student baseline across the three benchmarks, with no consistent benefit from increasing (Figure~\ref{fig:topk-ablation}).
Loss trajectories are similar across $K$ and plateau around optimizer step 20, with a comparable trend on internal data (Figure~\ref{fig:topk-support}(\subref{fig:topk-support-a})).
To balance performance and the storage and communication burden, we therefore use $K=128$ by default for this pair.

% The wider text block shortens this paragraph, so the next paragraph also wraps
% around the ablation figure; the full-width diagnostics follow both paragraphs.
\textbf{Does a compact support lose teacher information?}
For the $K=128$ ScaleSWE run, we examine the top-\(n\) token subsets selected by the student, where \(n\in\{4,8,64,128\}\) (Figure~\ref{fig:topk-support}(\subref{fig:topk-support-c},\subref{fig:topk-support-d})).
Averaged over steps 190--199, the student's top-\(4\), top-\(64\), and top-\(128\) capture \(98.4\%\), \(99.95\%\), and \(99.97\%\) of the \emph{teacher} probability mass, respectively; all coverage numbers in this paper are teacher mass measured on student-selected supports, not student mass.
Yet teacher--student top-$128$ overlap averages only $62.1\%$ over the same interval (Figure~\ref{fig:topk-support}(\subref{fig:topk-support-d})).
This overlap increases modestly early in training and then plateaus.
For $n\in\{4,8,64\}$,  overlap is computed using teacher rankings restricted to the top-$128$ candidates selected by the student; for $n=128$, teacher rankings span the full vocabulary.
Teacher candidates outside the student-selected support thus carry little probability mass on average. Incomplete token-set overlap need not imply substantial loss of teacher probability mass, so overlap and coverage should be treated as distinct quantities.
Together with the benchmark gains, these diagnostics suggest that compact student-selected candidate sets can support effective distillation in the evaluated setting.
\par
% End the ablation's text wrapping before the full-width diagnostics.
\WFclear
% Four 4:3 diagnostic panels share one figure and one caption.
\begin{figure}[!htb]
\centering
% Coordinates are percentages of the 5.5-inch canvas width in plot_row_layout.py.
% Anchor above each plot so a reference jump shows both the plot and its title below.
\begin{overpic}[width=\linewidth]{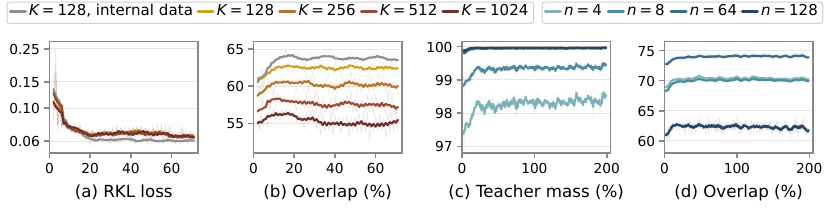}
\put(6,21.5){\phantomsubcaption\phantomsection\label{fig:topk-support-a}}
\put(30.7273,21.5){\phantomsubcaption\phantomsection\label{fig:topk-support-b}}
\put(56,21.5){\phantomsubcaption\phantomsection\label{fig:topk-support-c}}
\put(80.5455,21.5){\phantomsubcaption\phantomsection\label{fig:topk-support-d}}
\end{overpic}
\captionsetup{font=footnotesize,justification=raggedright,singlelinecheck=false}
\caption{\textbf{Top-$K$ selection and nested support diagnostics.} Qwen3.5-9B distilled from Qwen3.6-27B: (a,b) support-size sweep; (c,d) nested readouts for the $K=128$ ScaleSWE run. All horizontal axes show training steps. In (b), mean overlap over steps 62--71 is $62.5\%$ at $K=128$ and $55.0\%$ at $K=1024$.}
\label{fig:topk-support}
\label{fig:topk-support-k}
\label{fig:topk-support-n}
\end{figure}
% \FloatBarrier
\endgroup

\FloatBarrier
\section{Offline SWE-OPD}
\label{sec:offline}

Online SWE-OPD updates both the student parameters and the interaction histories used for subsequent supervision. Inspired by Lightning OPD~\citep{wu2026lightning}, Offline SWE-OPD freezes the latter by collecting trajectories from the initial student before optimization, and by scoring them with the teacher once (Figure~\ref{fig:main-pipeline} Right).  This design eliminates repeated teacher scoring to decouple student-teacher interaction. 
However, this also introduces a mismatch between the evolving student and the fixed training histories. We therefore treat offline training as an approximation to be evaluated. 

\subsection{Offline Supervision Construction}

We collect trajectories over the training set using the fixed initial policy $\pi_{\theta_0}$, following the repository interaction process in Section~\ref{sec:rollout}. 
For each trajectory, we save the task identifier, token sequence $\mathbf y$, and generation mask $\mathbf m$ defined in Section~\ref{sec:teacher}.
For ST training, we also save the initial student's log-probability $\log\pi_{\theta_0}(y_i\mid x,\mathbf y_{<i})$ at each generated position.

The frozen teacher evaluates $q_i$ on each saved student-generated prefix, as defined in Eq.~\ref{eq:shared-prefix}.
To avoid storing full-vocabulary values, we retain the teacher top-$K_{\mathrm{cache}}$ support
\begin{equation}
\mathcal{V}_{T,K_{\mathrm{cache}}}^{(i)}
=\operatorname{TopK}\!\left(q_i,K_{\mathrm{cache}}\right),
\qquad K_{\mathrm{cache}}=1024.
\label{eq:offline-buffer}
\end{equation}
where $\mathcal{V}_{T,K_{\mathrm{cache}}}^{(i)}$ denotes a teacher-defined token support, not a score.
For each supervised position, the offline buffer stores the token IDs and original log-probabilities $\log q_i(v)$ on this support. For example, $q_{i,(K_{\mathrm{cache}})}$ denotes the smallest teacher probability in the cache. 
Teacher scoring is performed once, and the resulting cache can be reused across multiple student-training epochs.

\subsection{Student Optimization on Fixed Prefixes}
\label{sec:offline-optimization}

During optimization, we evaluate the current student $p_i$ on the saved prefixes and apply the objectives from Section~\ref{sec:objective} using cached teacher scores.
For any token queried by either objective, RKL or ST, we use the following teacher-score lookup:
\begin{equation}
\widetilde q_i(v)=
\begin{cases}
q_i(v), & v\in\mathcal{V}_{T,K_{\mathrm{cache}}}^{(i)},\\
q_{i,(K_{\mathrm{cache}})}, & v\notin\mathcal{V}_{T,K_{\mathrm{cache}}}^{(i)}.
\end{cases}
\label{eq:offline-teacher-lookup}
\end{equation}
If a token in the student's current top-$K$ is absent from the teacher's cached top-$K_{\mathrm{cache}}$, we assign it the probability of the $K_{\mathrm{cache}}$-th token. Every omitted token satisfies $q_i(v)\leq q_{i,(K_{\mathrm{cache}})}$, so the fallback is an upper bound on its unknown teacher probability and avoids additional teacher queries~\footnote{A preliminary comparison with larger $K_{\mathrm{cache}}$ such as $4096$ yielded similar performance. Thus, to reduce the burden of storage, we use 1024 by default.}.

\emph{RKL Loss.}
For RKL training, we recompute the student top-$K$ support $\mathcal{V}_{K}^{(i)}$ in Eq.~\ref{eq:student-topk} at each update, with $K=128$.
In other words, the student support remains dynamic rather than being frozen to the initial student $\pi_{\theta_0}$. 
We substitute $\widetilde q_i$ for $q_i$ in Eq.~\ref{eq:rkl-topk}, retaining the support-normalized student weights $\bar p_i$ with the same stop-gradient on their normalization constant.

\emph{ST Loss.}
For ST training, we use Eq.~\ref{eq:sampled-token-loss} with $q_i(y_i)$ replaced by $\widetilde q_i(y_i)$ and the old policy fixed to the initial student, $p_i^{\mathrm{old}}(v)=\pi_{\theta_0}(v\mid x,\mathbf y_{<i})$ so as to be consistent with the fixed trajectories. Although the teacher score is fixed, the advantage changes with the current student:
\begin{equation}
\ell_i^{\mathrm{ST}} =
-\frac{p_i(y_i)}{\pi_{\theta_0}(y_i\mid x,\mathbf{y}_{<i})}
\operatorname{sg}\!\left[\log\widetilde q_i(y_i)-\log p_i(y_i)\right] \, .
\label{eq:offline-sampled-token-loss}
\end{equation}
Both objectives use the aggregation in Eq.~\ref{eq:swe-opd-objective}. The saved trajectories and cached scores are constants, and only the current student parameters $\theta$ are updated. This allows both forms of distillation to reuse offline supervision across optimization steps.

\emph{Cross-tokenizer ST Loss.}
\label{sec:cross-tokenizer}
In real applications, it is common for the student and teacher to have different tokenizers.
To validate our method under such a scenario, we further adapt offline ST to a teacher with a different tokenizer in a naive way:
we re-encode each saved trajectory with the teacher tokenizer and call a generated position \emph{aligned} if some teacher token covers exactly the same characters as $y_i$.  Unaligned generated tokens are retained in the context and merged using the simplest GOLD method, with their log-probabilities represented by an average.  Note that  we deliberately opt for this simple and conservative cross-tokenizer ST adaptation, avoiding sophisticated cross-vocabulary mapping mechanisms or modifications to the original distillation objective. Although more advanced cross-vocabulary mapping methods~\citep{niu2026breakingtokenizerbarrieronpolicy, sun2026simctrecoveringlostsupervision, he2026simpleopdsimpletokenizeragnosticonpolicy} are possible, we leave them as future work.

\subsection{Performance and Cost of Offline Supervision}
\label{sec:offline-results}

\textbf{Can fixed initial-student trajectories still support effective distillation?} We compare online and offline SWE-OPD trained on the
same internal data, with matched student-training budgets (Tab.~\ref{tab:offline-main}).
All evaluated offline configurations improve over
their student baselines on all three benchmarks, so freezing the interaction
histories does not prevent effective learning within this budget.

\begin{table}[htbp]
\centering
\caption{Results of SWE-OPD and Offline SWE-OPD on the internal dataset. We compare both variants for Qwen3.5-9B distilled from Qwen3.6-27B and GLM-4.5-Air distilled from GLM-5.2, and additionally evaluate Offline SWE-OPD (ST) for GLM-4.5-Air distilled from DeepSeek-V4-Flash-0731, a teacher with a different tokenizer.}
\label{tab:offline-main}
\small
\begin{tabular*}{\linewidth}{@{\extracolsep{\fill}}lcccccc}
\toprule
\multirow{2}{*}{Method}
& \multicolumn{2}{c}{Verified}
& \multicolumn{2}{c}{Pro}
& \multicolumn{2}{c}{Multilingual} \\
\cmidrule(lr){2-3}\cmidrule(lr){4-5}\cmidrule(lr){6-7}
& Score & $\Delta$ & Score & $\Delta$ & Score & $\Delta$ \\
\midrule
\multicolumn{7}{l}{\emph{Student: Qwen3.5-9B \quad Teacher: Qwen3.6-27B}} \\
\quad Student baseline
    & 57.9 &  & 35.9 &  & 43.0 &  \\
\quad SWE-OPD (RKL)
    & {61.6} & $+3.7$ & 38.4 & $+2.5$ & 44.3 & $+1.3$ \\
\quad SWE-OPD (ST)
    & 61.3 & $+3.4$ & \textbf{38.7} & $+2.8$ & 43.1 & $+0.1$ \\
\quad Offline SWE-OPD (RKL)
    & 60.2 & $+2.3$ & 38.0 & $+2.1$ & {45.3} & $+2.3$ \\
\quad Offline SWE-OPD (ST)
    & \textbf{63.7} & $+5.8$ & 36.5 & $+0.6$ & \textbf{46.0} & $+3.0$ \\
\midrule
\multicolumn{7}{l}{\emph{Student: GLM-4.5-Air \quad Teacher: GLM-5.2}} \\
\quad Student baseline
    & 59.9 &  & 35.8 &  & 41.6 &  \\
\quad SWE-OPD (ST)
    & \textbf{60.4} & $+0.5$ & 37.1 & $+1.3$ & \textbf{45.0} & $+3.4$ \\
\quad Offline SWE-OPD (ST)
    & \textbf{60.4} & $+0.5$ & \textbf{37.5} & $+1.7$ & {44.6} & $+3.0$ \\
\midrule
\multicolumn{7}{l}{\emph{Student: GLM-4.5-Air \quad Teacher: DeepSeek-V4-Flash-0731}} \\
% Cross-tokenizer run (Section~\ref{sec:cross-tokenizer}); deltas use the GLM-4.5-Air baseline above.
% \quad \modified{Offline SWE-OPD (Top RKL yan)}
%     & \textbf{61.5} & $+1.6$ & \textbf{42.7} & $+6.9$ & \textbf{46.7} & $+5.1$ \\
\quad Offline SWE-OPD (ST)
    & \textbf{62.5} & $+2.6$ & \textbf{39.7} & $+3.9$ & \textbf{46.2} & $+4.6$ \\
\bottomrule
\end{tabular*}
\end{table}

For Qwen, offline ST achieves the highest Verified
and Multilingual scores among the Qwen
configurations.
Relative to online ST, it gains 2.4\% and 2.9\%
on Verified and Multilingual, respectively, while scoring 2.2\% lower on Pro.
Offline RKL also gains 1.0 point on Multilingual while
scoring 1.4\% and 0.4\% lower on Verified and Pro than its online counterpart.
For GLM-4.5-Air, offline ST matches the reported online
score on Verified, gains 0.4\% on Pro, and scores 0.4\% lower on Multilingual.
Together with the resource savings in
Section~\ref{sec:offline-efficiency}, these results
make cached supervision a promising option for
enhancing large students under limited training resources.

\textbf{Does offline ST transfer across tokenizers?} We distill GLM-4.5-Air from DeepSeek-V4-Flash-0731, whose tokenizer differs from the student's, using the naive cross-tokenizer ST variant.
We evaluate only offline training because, with a shared tokenizer, offline ST stays within 0.4\% of online ST for GLM-4.5-Air on every benchmark at fewer GPU-hours (Section~\ref{sec:offline-efficiency}).
Table~\ref{tab:offline-main} shows that even with a simple mapping strategy across different tokenizers, distillation from a powerful teacher yields substantial gains across the GLM-4.5-Air configurations. Specifically, it improves GLM-4.5-Air by 2.6\%, 3.9\%, and 4.6\% on Verified, Pro, and Multilingual. The results imply that enhancing distillation performance more effectively is a promising direction.

% This suggests that replaying fixed trajectories for additional epochs can strengthen learning on some task distributions, but does not replace the broader state coverage obtained from fresh online rollouts. We therefore view the offline variant primarily as a resource-efficient alternative rather than a strict replacement for online SWE-OPD.
\subsection{Resource Cost and Its Attribution}
\label{sec:offline-efficiency}

\begin{wraptable}{r}{0.46\columnwidth}
\centering
\caption{Training cost on the internal dataset (GPU-hours) for our online and offline implementations.}
\label{tab:offline-efficiency}
\small
\setlength{\tabcolsep}{3pt}
\resizebox{\linewidth}{!}{%
\begin{tabular}{lcc}
\toprule
\textbf{Method}
    & \textit{Qwen3.5-9B}
    & \textit{GLM-4.5-Air} \\
\midrule
SWE-OPD
    & 1024 & 2808 \\
Offline SWE-OPD
    & 416  & 768 \\
\midrule
GPU-hour reduction
    & \textbf{2.46$\times$}
    & \textbf{3.65$\times$} \\
\midrule
\multicolumn{3}{l}{\textit{Offline SWE-OPD breakdown}} \\
Student rollout
    & 176 & 368 \\
Teacher scoring
    & 48  & 160 \\
OPD training
    & 192 & 240 \\
\bottomrule
\end{tabular}%
}
\end{wraptable}

\textbf{What do the cost differences measure?} Relative to our online implementation, offline execution reduces end-to-end GPU-hours from 1024 to 416 (59.4\%) for Qwen3.5-9B and from 2808 to 768 (72.6\%) for GLM-4.5-Air, i.e.\ reduction factors of 2.46$\times$ and 3.65$\times$ in Table~\ref{tab:offline-efficiency}. Reusing fixed
student trajectories and cached teacher scores eliminates further rollout collection and teacher inference during offline optimization.

More importantly, offline execution removes the need to run all stages concurrently.
Sequential execution lets the Offline SWE-OPD stages reuse one GPU pool,
reducing peak allocation to $\max(N_R,N_T,N_S)$, where the three terms
denote GPU allocations for rollout, teacher scoring,
and student optimization.
In our setup, this halves the peak GPU requirement
from 64 to 32 GPUs, making SWE-OPD feasible on substantially smaller clusters.
Table~\ref{tab:offline-efficiency} reports the end-to-end GPU-hours of SWE-OPD together with a breakdown of the offline cost. 
All measurements are collected on the same internal dataset and hardware, including the cost of failed or retried rollouts.

% Finish the preceding wraptable before the next section.
\WFclear
\section{Diagnosing Training Collapse under the top-k reverse-KL Loss}
\label{sec:stability}
\label{sec:why-topk-fails}

The RKL loss improves the Qwen student, yet GLM-4.5-Air develops severe repetition while ST
training remains stable. We examine how the RKL loss converts teacher scores into updates and
identify support escape as a potential cause. We then connect this mechanism to the training dynamics observed in the GLM setting.

\subsection{How GLM Training Collapses}
\label{sec:loss-mismatch}

\begin{figure}[!htbp]
\centering
\captionsetup{font=footnotesize}
\captionsetup[subfigure]{font=small}
\includegraphics[width=\linewidth]{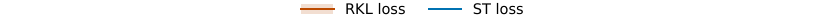}\par
\begin{subfigure}[t]{0.328\linewidth}
  \centering
  \begin{overpic}[width=\linewidth]{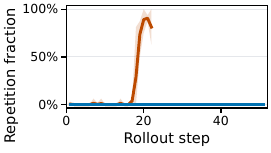}
    \put(96,44){\makebox[0pt][r]{\footnotesize(a)}}
  \end{overpic}
  \phantomsubcaption
  \label{fig:wf-training-repetition}
\end{subfigure}\hfill%
\begin{subfigure}[t]{0.328\linewidth}
  \centering
  \begin{overpic}[width=\linewidth]{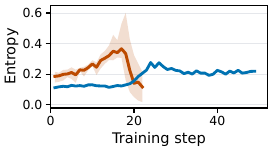}
    \put(96,44){\makebox[0pt][r]{\footnotesize(b)}}
  \end{overpic}
  \phantomsubcaption
  \label{fig:wf-training-entropy}
\end{subfigure}\hfill%
\begin{subfigure}[t]{0.328\linewidth}
  \centering
  \begin{overpic}[width=\linewidth]{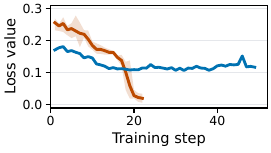}
    \put(96,44){\makebox[0pt][r]{\footnotesize(c)}}
  \end{overpic}
  \phantomsubcaption
  \label{fig:wf-training-loss}
\end{subfigure}
\caption{GLM training dynamics:
(\subref{fig:wf-training-repetition}) Repetition fraction;
(\subref{fig:wf-training-entropy}) student entropy;
(\subref{fig:wf-training-loss}) training loss.
RKL and ST training are shown in orange and blue, respectively.}
\label{fig:wf-training}
\end{figure}

Figure~\ref{fig:wf-training} shows a pronounced separation between loss reduction and useful behavior: repetition rises sharply as the student entropy falls after an initial increase. The loss decreases overall as repetition rises. The model therefore becomes concentrated on repetitive continuations while still improving its training objective. A small distillation loss is insufficient evidence that teacher supervision is improving the student's behavior.
In comparison, ST training records no repetition and avoids the sharp terminal entropy decline. Together with its benchmark gains in Table~\ref{tab:internal}, this contrast motivates comparing how the two objectives turn teacher scores into student updates.

\subsection{Support Escape in the Truncated Loss}
\label{sec:support-escape}
\label{sec:wf-mechanism}

We now examine what kind of update Eq.~\ref{eq:rkl-topk} performs. Let $\hat{\Zeta}_{\mathrm{total}}=\sum_{v\in\mathcal V_K^{(i)}}p_i(v)$ denote the student's total probability mass on the selected top-$K$ support.
Differentiating Eq.~\ref{eq:rkl-topk} with respect to the student parameters $\theta$ at a fixed prefix and support, gives (see Appendix~\ref{app:wf-forward-gradient} for the full derivation):
\begin{equation}
\nabla_\theta\ell_i^{\mathrm{RKL}}
=\underbrace{\sum\nolimits_{v\in\mathcal V_K^{(i)}}\bar p_i(v)
\log\tfrac{p_i(v)}{q_i(v)}\nabla_\theta\log p_i(v)}_{\mathbf g_i:\;\text{teacher-dependent supervision}}
+\underbrace{\vphantom{\sum\nolimits_{v\in\mathcal V_K^{(i)}}\bar p_i(v)\log\tfrac{p_i(v)}{q_i(v)}\nabla_\theta\log p_i(v)}
\nabla_\theta\log\hat{\Zeta}_{\mathrm{total}}}_{\mathbf h_i:\;\text{support-mass gradient}}.
\label{eq:wf-gradient-split}
\end{equation}
% Start the wrap after the equation so the surrounding prose fills its height.
\begin{wrapfigure}{r}{0.3\textwidth}
\centering
\includegraphics[width=\linewidth]{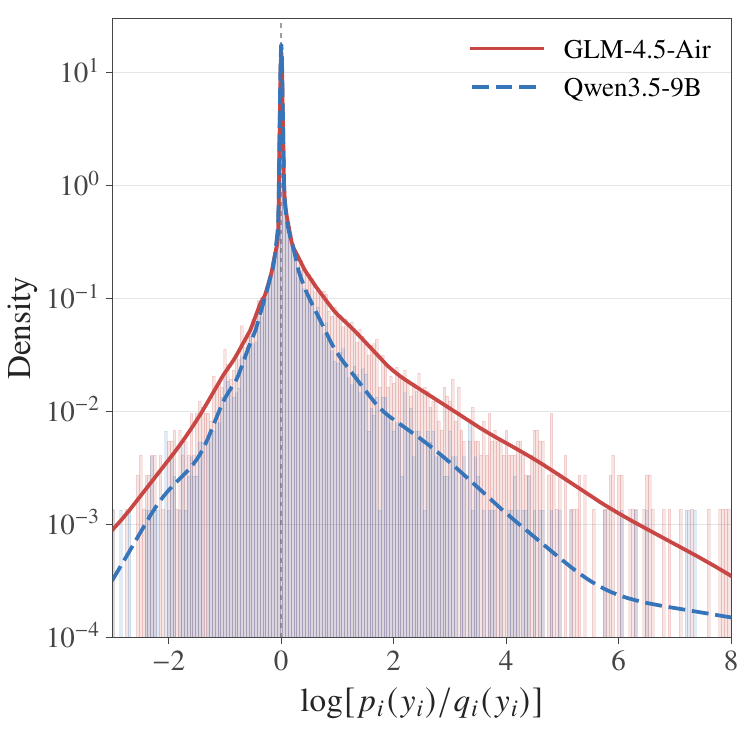}
\captionsetup{font=footnotesize,justification=raggedright,singlelinecheck=false}
\caption{{teacher-student probability divergence.}}
\label{fig:wf-logratio}
\end{wrapfigure}

Here $\mathbf g_i$ carries teacher-dependent supervision, while $\mathbf h_i$ is the gradient of the log of the retained probability mass. A sufficiently small step along $-\mathbf h_i$ alone decreases $\hat{\Zeta}_{\mathrm{total}}$ whenever $\mathbf h_i\ne0$, shifting probability outside the scored support. We call this decrease in retained mass \emph{support escape}; it occurs under the full update when $\mathbf g_i$ does not offset this outward tendency. Even at teacher agreement ($p_i=q_i$) on the retained support, $\mathbf g_i=0$ while $\mathbf h_i$ may remain nonzero, so the update need not be stationary at agreement. % Over the full vocabulary $\hat{\Zeta}_{\mathrm{total}}=1$ and $\mathbf h_i=0$, so the term is introduced by truncation.

% 顺着上面的推导就能看出问题：如果直接在 top-k 上训练 KL in Loss，优化会持续减小 
% ，把概率质量不断从 top-k 内部推向 tail。但 tail 里的 token 完全不在监督范围内——学生可以毫无代价地把任意概率堆到那里，不会因此受到任何惩罚。结果就是分布逐渐失控、训练崩溃。

\paragraph{Support Escape and GLM Collapse.}
Support escape reveals a structural failure mode of the top-K RKL loss.
It pushes probability mass from the target $\mathcal{V}_K$
set into the tail, which lack
the token-specific teacher correction, thus assigning
mass to candidates the teacher considers unlikely.
Repeated updates can therefore reduce the loss while
moving the student away from the teacher.
In our training, these shifts alter subsequent actions
and supervision prefixes, carrying the misalignment into future
rollouts.
This may explain how GLM's loss can continue to decrease as its
generations become increasingly repetitive and eventually collapse.
We also provide detailed analysis with a mechanism example in Appendix~\ref{app:wf-derivation}.

\paragraph{Why the two students Qwen and GLM may differ.}
Figure~\ref{fig:wf-logratio} shows a heavier positive tail of
sampled-action log-ratios for GLM, whereas Qwen is more
concentrated near zero.
GLM thus more frequently selects actions to which its teacher
assigns substantially lower probability.
Under the RKL loss, suppressing such actions within the
retained support can redirect probability toward unscored tail
tokens. We speculate that these stronger conflicts amplify support escape and drive rollout degeneration
in GLM, whereas Qwen’s closer teacher–student agreement limits this additional pressure and
allows effective distillation.
This provides a reference for designing efficient objectives that preserve
agreement while controlling probability mass outside the scored support.
\par
\WFclear

\section{Conclusion}

% We studied vanilla on-policy distillation for agentic coding. Teacher supervision on student-visited prefixes transfers coding capability across model scales. And our offline variant retains gains over the student baselines at 59.4\%--72.6\% fewer GPU-hours and half the peak GPU allocation, with benchmark-dependent accuracy. We also analyze the GLM failure potentially due to the probability leak to unscored tail tokens though the top-$K$ reverse-KL loss keeps decreasing. These results provide a baseline for studying how teacher supervision can complement outcome rewards and selective data curation. 

In this paper, we studied on-policy distillation for SWE agents. Vanilla SWE-OPD trains the student with teacher supervision on the prefixes that the student itself visits without filtering trajectories by task success. We compare and analyze two forms of this supervision: a top-$K$ RKL loss over the top-$K$ candidates of the student, and an ST loss that uses only the teacher score of the sampled token. Extensive experiments demonstrate that on-policy supervision transfers coding capability. SWE-OPD improves Qwen3.5-9B and GLM-4.5-Air on SWE-bench Verified, Pro, and Multilingual.

Offline SWE-OPD further indicates that these prefixes do not need to be refreshed during training. It scores the trajectories of the initial student with the teacher only once, yet it still improves over the student baselines. Moreover, it reduces end-to-end GPU-hours by 59.4\%--72.6\% while halving peak GPU allocation. The same pipeline also supports a teacher with a different tokenizer. Distilling GLM-4.5-Air from DeepSeek-V4-Flash-0731 shows that OPD can transfer coding capability across model families.

We also find that a decreasing loss does not necessarily indicate improved agent performance. Under the top-$K$ RKL loss, GLM-4.5-Air exhibits severe repetition despite a steadily decreasing loss. Our gradient analysis identifies such failure as potentially due to the probability leak to unscored tail tokens.

Overall, both online and offline SWE-OPD improve SWE agents. Offline SWE-OPD achieves these gains at a substantially lower compute cost. The repetition failure under the top-$K$ RKL loss also calls for reconsidering the design of OPD objectives. These results provide a baseline for studying how teacher supervision can complement outcome rewards and selective data curation.

\section{Future Work}

Given our conservative design choices, we believe there remains substantial room for further improvement as listed below.

First, as we have mentioned before, we can develop more advanced variants of OPD to further improve the student’s performance. In addition, OPD can be combined with reinforcement learning to jointly leverage process-level and outcome-level rewards~\citep{yang2026self, yang2026learning}, potentially
providing more effective and comprehensive supervision.

Second, training data plays a critical role in OPD. To further improve performance, it is important
to investigate which types of data are most suitable for OPD.

Third, we only evaluate the effectiveness of OPD on agentic coding using models of different sizes
within the same model family, which essentially share the same tokenizer. A natural next step is
therefore to investigate whether OPD remains effective across different model families with distinct
tokenizers~\citep{wang2026cross, sun2026simct, pati2025, niu2026breaking}. In this work, we conduct a preliminary experiment using DeepSeek-V4-Flash-0731 and GLM-4.5-Air and verify the feasibility.

Last but not least, OPD is currently primarily applied to post-training, e.g., capability merging. Since
OPD essentially transfers knowledge from a teacher model to a student model, it is worth exploring
whether OPD can also be extended to pretraining, potentially enabling knowledge
transfer at earlier stages of model development with less training cost.

% \subsection*{AI use statement}

% We used generative AI tools to find related work on on-policy distillation and agentic coding, and to draft and polish descriptions of the SWE-OPD and Offline SWE-OPD pipelines. AI also assisted with deriving and verifying mathematical claims about the RKL loss, ST gradients, and offline teacher-score approximation, and with checking that probability normalization and loss masking were consistent between the formulas and code. The authors reviewed all AI-assisted material and take full responsibility for the paper.

\clearpage
\bibliography{iclr2027_conference}
\bibliographystyle{colm2024_conference}

\clearpage
\appendix
\section{Experimental Setup and Implementation Details}
\label{app:implementation}

\subsection{Training Data and Model Configurations}
\label{app:training-configurations}

\paragraph{Training subsets.}
We randomly sample 5k queries from the 20k-query ScaleSWE dataset~\citep{zhao2026immersiongithubuniversescaling} and approximately 4k tasks from the internal dataset~\citep{guo2026menvagent}. These subset sizes align the training-data budget with SWE reinforcement learning settings that typically use a few thousand tasks~\citep{jain2025r2e,uniagent_github}. We use the same ScaleSWE subset for Teacher Rollout Distillation and SWE-OPD, and the same internal subset for the online and offline comparisons. In preliminary experiments, training on other randomly sampled subsets also produced clear improvements over the corresponding student baselines.

\paragraph{Models and objectives.}
We use instruction-tuned checkpoints for all model pairs: Qwen3.5-9B distilled from Qwen3.6-27B, GLM-4.5-Air distilled from GLM-5.2, and GLM-4.5-Air distilled from DeepSeek-V4-Flash-0731. The last pair uses different tokenizers and is trained only with offline ST (Section~\ref{sec:cross-tokenizer}). Table~\ref{tab:experiment-configurations} summarizes the training configurations. For RKL training, the student support is recomputed at each update with $K=128$ for the Qwen and GLM pairs; the support-size ablation additionally evaluates $K\in\{256,512,1024\}$ on ScaleSWE. ST uses the generated token and does not have a top-$K$ training support. All offline configurations use a teacher cache of $K_{\mathrm{cache}}=1024$, as described in Section~\ref{sec:offline}. GLM RKL training collapses and is analyzed separately in Section~\ref{sec:why-topk-fails}.

\begin{table}[!htbp]
\centering
\caption{Training configurations. Qwen denotes Qwen3.5-9B $\leftarrow$ Qwen3.6-27B, GLM denotes GLM-4.5-Air $\leftarrow$ GLM-5.2, and GLM/DS denotes GLM-4.5-Air $\leftarrow$ DeepSeek-V4-Flash-0731, with student $\leftarrow$ teacher. $K$ is the RKL student-support size; $K_{\mathrm{cache}}$ is the offline teacher-cache size. A dash denotes a setting that does not apply.}
\label{tab:experiment-configurations}
\small
\setlength{\tabcolsep}{4pt}
\begin{tabular*}{\linewidth}{@{\extracolsep{\fill}}ll l c c c@{}}
\toprule
Training data & Model pair & Training variant & Loss & $K$ & $K_{\mathrm{cache}}$ \\
\midrule
ScaleSWE (5K) & Qwen & Teacher rollouts & SFT & -- & -- \\
 & Qwen & Online & RKL & 128 & -- \\
 & Qwen & Online & ST & -- & -- \\
 & Qwen & Online (ablation) & RKL & 256/512/1024 & -- \\
\midrule
Internal ($\sim$4K) & Qwen & Online & RKL & 128 & -- \\
 & Qwen & Online & ST & -- & -- \\
 & Qwen & Offline & RKL & 128 & 1024 \\
 & Qwen & Offline & ST & -- & 1024 \\
 & GLM & Online$^\dagger$ & RKL & 128 & -- \\
 & GLM & Online & ST & -- & -- \\
 & GLM & Offline & ST & -- & 1024 \\
 & GLM/DS & Offline & ST & -- & 1024 \\
\bottomrule
\end{tabular*}
\par\vspace{3pt}
\begin{minipage}{\linewidth}
\footnotesize $^\dagger$ GLM RKL training collapses and has no reported benchmark score; see Section~\ref{sec:why-topk-fails}.
\end{minipage}
\end{table}

\subsection{Optimization and Checkpoint Selection}
\label{app:optimization}

We implement OPD using Slime~\citep{slime_github} and optimize the student with AdamW at a constant learning rate of $10^{-6}$. For ST training, we clip the importance ratio to $[0.8,1.2]$. The rollout batch size is 64, with one rollout sampled per training query. Training rollouts use a 128K-token context budget. All training configurations use a one-epoch budget over their training subset, and online and offline SWE-OPD use matched student-training budgets on the same sampled tasks. The online variant refreshes student trajectories during training; the offline variant trains on the initial student's saved trajectories and cached teacher scores.

For SWE-OPD and Offline SWE-OPD, we select the evaluated checkpoint within 1.0 epochs. Within this interval, we choose the checkpoint with the highest teacher--student top-$K$ overlap, using the diagnostic in Section~\ref{sec:topk-analysis}.
In particular, most checkpoints are selected at 1 epoch; for GLM-4.5-Air, the online SWE-OPD checkpoint is selected at around half an epoch.
This selection rule applies to both RKL and ST training, including the offline configurations. In preliminary experiments, extending training beyond one epoch provided little additional benefit and sometimes reduced downstream performance.

\paragraph{Teacher Rollout Distillation.}
For the SFT baseline in Table~\ref{tab:main}, Qwen3.6-27B generates approximately 5,000 trajectories on the same ScaleSWE subset. We retain successful and unsuccessful trajectories without filtering by task reward, and fine-tune Qwen3.5-9B on the teacher-generated response tokens for one epoch using the standard SFT objective.

\subsection{Agent Scaffold and Tools}
\label{app:agent-scaffold}

We adopt the ScaleSWE scaffold~\citep{zhao2026immersiongithubuniversescaling} through its function-calling configuration (\texttt{edit\_fn\_calling}) in our R2E-Gym integration~\citep{jain2025r2e}. Training and evaluation share the scaffold loader and prompt-construction code. Each task starts with a system message and a user message containing the issue description and repository location. Tool schemas are supplied through the model's chat template, and observations are returned as tool-role messages. The agent has access to three tools:

\begin{itemize}[leftmargin=*,itemsep=2pt,topsep=4pt]
\item \texttt{str\_replace\_editor}: view files or directories, create files, replace an exact text span, and insert text after a specified line. File changes persist across calls.
\item \texttt{execute\_bash}: execute a shell command in the repository, including search, build, and test commands. Each call starts a fresh shell; directory changes and shell environment variables do not persist between calls. Commands are subject to a timeout, and long outputs may be truncated.
\item \texttt{finish}: terminate the interaction when the task is complete or the agent cannot proceed further. The tool takes no arguments.
\end{itemize}

\subsection{Evaluation Protocol}
\label{app:evaluation}

We follow the R2E-Gym evaluation protocol~\citep{jain2025r2e}, using SGLang~\citep{zheng2024sglang} to serve the evaluated models. The agent interacts with the task repository using the scaffold above, and its resulting patch is evaluated using the benchmark's required functionality and regression tests. A task is resolved only if all required tests pass.

We evaluate on 500 SWE-bench Verified tasks, 731 SWE-bench Pro tasks, and 300 SWE-bench Multilingual tasks. The evaluation temperature is 1.0, top-$p$ is 0.95, and the repetition penalty is 1.0. We allow at most 100 interaction steps on Verified and 150 on Pro and Multilingual, with a 128K-token context budget. These evaluation settings apply to the student baselines and the trained checkpoints in the benchmark tables.

\subsection{System and User-Message Templates}
\label{app:prompt-templates}

The colored blocks reproduce the system and user-message templates from the ScaleSWE \texttt{edit\_fn\_calling} configuration verbatim. The placeholder \texttt{\{problem\_statement\}} is filled with the task's issue description; \texttt{/testbed} is replaced by the actual repository working directory when it differs. The user-message template retains its original wording, including \emph{python code repository}. The three tool schemas described in Appendix~\ref{app:agent-scaffold} are supplied separately through the model's chat template.

\begin{agentprompt}{promptblue}{ScaleSWE scaffold: system message}
\begin{Verbatim}[fontsize=\footnotesize,baselinestretch=0.95,breaklines=true,breakanywhere=true,breaksymbolleft={},breaksymbolright={}]
You are a helpful AI assistant that can interact with a computer to solve tasks.

<ROLE>
Your primary role is to assist by executing commands, modifying code, and solving technical problems effectively. You should be thorough, methodical, and prioritize quality over speed.
</ROLE>

<EFFICIENCY>
* Each action you take is somewhat expensive. Wherever possible, combine multiple actions into a single action, e.g. combine multiple bash commands into one, using sed and grep to edit/view multiple files at once.
* When exploring the codebase, use efficient tools like find, grep, and git commands with appropriate filters to minimize unnecessary operations.
</EFFICIENCY>

<FILE_SYSTEM_GUIDELINES>
* When given a file path, do NOT assume it's relative to the current working directory. First explore the file system to locate the file before working on it.
* If asked to edit a file, edit the file directly, rather than creating a new file with a different filename.
* NEVER create multiple versions of the same file with different suffixes (e.g., file_test.txt, file_fix.txt, file_simple.txt). Instead:
- Always modify the original file directly when making changes
- If you need to create a temporary file for testing, delete it once you've confirmed your solution works
- If you decide a file you created is no longer useful, delete it instead of creating a new version
* When reproducing bugs or implementing fixes, use a single file rather than creating multiple files with different versions
</FILE_SYSTEM_GUIDELINES>

<CODE_QUALITY>
* Write clean, efficient code with minimal comments. Avoid redundancy in comments: Do not repeat information that can be easily inferred from the code itself.
* When implementing solutions, focus on making the minimal changes needed to solve the problem.
* Before implementing any changes, first thoroughly understand the codebase through exploration.
* If you are adding a lot of code to a function or file, consider splitting the function or file into smaller pieces when appropriate.
* Place all imports at the top of the file unless placing imports at the top would cause issues (e.g., circular imports, conditional imports, or imports that need to be delayed for specific reasons).
</CODE_QUALITY>

<PROBLEM_SOLVING_WORKFLOW>
1. EXPLORATION: Thoroughly explore relevant files and understand the context before proposing solutions
2. ANALYSIS: Consider multiple approaches and select the most promising one
3. TESTING:
  * For bug fixes: Write a small reproduction script to confirm the issue exists before implementing a fix, and re-run it afterward to verify the fix works.
  * For new features: Consider writing simple tests when appropriate to validate your implementation.
4. IMPLEMENTATION:
  * Make focused, minimal changes to address the problem
  * Always modify existing files directly rather than creating new versions with different suffixes
  * If you create temporary files for testing, delete them after confirming your solution works
5. VERIFICATION: Test your implementation thoroughly, including edge cases.
6. SUBMISSION: When you have completed the task or cannot proceed further, call the `finish` tool to end the interaction.
</PROBLEM_SOLVING_WORKFLOW>

<ENVIRONMENT_SETUP>
* The development environment is already set up and all dependencies are installed. You can directly run tests or compile without additional setup.
* If you do encounter a missing tool or package, you may install it as needed.
</ENVIRONMENT_SETUP>

<TROUBLESHOOTING>
* If you've made repeated attempts to solve a problem but tests still fail:
1. Step back and reflect on 5-7 different possible sources of the problem
2. Assess the likelihood of each possible cause
3. Methodically address the most likely causes, starting with the highest probability
* When you run into any major issue while executing your plan, do not try to directly work around it. Instead, re-evaluate your approach and consider alternative solutions.
</TROUBLESHOOTING>

<PROCESS_MANAGEMENT>
* Be aware that each bash command must complete and return within a time limit. Avoid running commands that block the terminal indefinitely (e.g., starting a server in the foreground, interactive prompts, or commands that wait for user input). If you need to start a long-running process, run it in the background.
* When terminating processes, use specific identifiers rather than broad patterns like `pkill -f python`. Prefer finding the exact PID with `ps aux` first.
</PROCESS_MANAGEMENT>

<TOOL_CALL_RULES>
* Only call one function at a time.
* You may provide reasoning in natural language BEFORE the function call.
* Required parameters MUST be specified.
</TOOL_CALL_RULES>
\end{Verbatim}
\end{agentprompt}

\begin{agentprompt}{promptteal}{ScaleSWE scaffold: user-message template}
\begin{Verbatim}[fontsize=\footnotesize,baselinestretch=0.95,breaklines=true,breakanywhere=true,breaksymbolleft={},breaksymbolright={}]
<uploaded_files>
/testbed
</uploaded_files>

I've uploaded a python code repository in the directory /testbed. Consider the following issue description:

<issue_description>
{problem_statement}
</issue_description>

Can you help me implement the necessary changes to the repository so that the requirements specified in the <issue_description> are met?
I've already taken care of all changes to any of the test files described in the <issue_description>. This means you DON'T have to modify the testing logic or any of the tests in any way!
Also the development environment is already set up for you (i.e., all dependencies already installed), so you don't need to install other packages.
Your task is to make the minimal changes to non-test files in the /testbed directory to ensure the <issue_description> is satisfied.

Follow these phases to resolve the issue:

Phase 1. READING: read the problem and reword it in clearer terms
  1.1 If there are code or config snippets. Express in words any best practices or conventions in them.
  1.2 Highlight message errors, method names, variables, file names, stack traces, and technical details.
  1.3 Explain the problem in clear terms.
  1.4 Enumerate the steps to reproduce the problem.
  1.5 Highlight any best practices to take into account when testing and fixing the issue

Phase 2. RUNNING: run the tests on the repository
  2.1 The development environment is already set up and all dependencies are installed -- you do NOT need to install anything.
  2.2 Explore the repository structure and figure out how to run the relevant tests.

Phase 3. EXPLORATION: find the files that are related to the problem and possible solutions
  3.1 Use `grep` to search for relevant methods, classes, keywords and error messages.
  3.2 Identify all files related to the problem statement.
  3.3 Propose the methods and files to fix the issue and explain why.
  3.4 From the possible file locations, select the most likely location to fix the issue.

Phase 4. TEST CREATION: before implementing any fix, create a script to reproduce and verify the issue.
  4.1 Look at existing test files in the repository to understand the test format/structure.
  4.2 Create a minimal reproduction script that reproduces the located issue.
  4.3 Run the reproduction script to confirm you are reproducing the issue.
  4.4 Adjust the reproduction script as necessary.

Phase 5. FIX ANALYSIS: state clearly the problem and how to fix it
  5.1 State clearly what the problem is.
  5.2 State clearly where the problem is located.
  5.3 State clearly how the test reproduces the issue.
  5.4 State clearly the best practices to take into account in the fix.
  5.5 State clearly how to fix the problem.

Phase 6. FIX IMPLEMENTATION: Edit the source code to implement your chosen solution.
  6.1 Make minimal, focused changes to fix the issue.

Phase 7. VERIFICATION: Test your implementation thoroughly.
  7.1 Run your reproduction script to verify the fix works.
  7.2 Add edge cases to your test script to ensure comprehensive coverage.
  7.3 Run existing tests related to the modified code to ensure you haven't broken anything.

Phase 8. FINAL REVIEW: Carefully re-read the problem description and compare your changes with the original code.
  8.1 Ensure you've fully addressed all requirements.
  8.2 Run any tests in the repository related to:
  8.2.1 The issue you are fixing
  8.2.2 The files you modified
  8.2.3 The functions you changed
  8.3 If any tests fail, revise your implementation until all tests pass

Phase 9. SUBMISSION: When you have completed the task or cannot proceed further, call the `finish` tool to end the interaction.

Be thorough in your exploration, testing, and reasoning. It's fine if your thinking process is lengthy - quality and completeness are more important than brevity.

--------------------- END OF NEW TASK DESCRIPTION ---------------------

PLEASE follow the format strictly! PLEASE EMIT ONE AND ONLY ONE FUNCTION CALL PER MESSAGE.
\end{Verbatim}
\end{agentprompt}

\section{Gradient Analysis of the Implemented Top-$K$ loss}
\label{app:wf-derivation}

\paragraph{Scope and notation.}
We follow Section~\ref{sec:objective}, including its omission of the batch index $b$. For each derivative, we fix token position $i$, its prefix $(x,\mathbf y_{<i})$, the teacher distribution $q_i$, and the selected support $\mathcal V_K^{(i)}$ from Eq.~\ref{eq:student-topk}. The student probabilities $p_i(v)$ from Eq.~\ref{eq:shared-prefix} are differentiable functions of $\theta$ and are positive unless an endpoint convention is stated. The support and the detached normalizer $\Zeta_i^p$ are recomputed between training updates. The loss decomposition uses parameter gradients; Appendix~\ref{app:wf-constructed-example} separately gives the logit derivatives used for the mechanism figure.

\subsection{Forward Value and Backward Computation}
\label{app:wf-forward-gradient}

\paragraph{Forward value and detached normalization.}
Using the support mass defined in Section~\ref{sec:wf-mechanism}, we distinguish this differentiable quantity from its detached copy used to normalize the loss:
\[
\hat{\Zeta}_{\mathrm{total}}
=\sum_{v\in\mathcal V_K^{(i)}}p_i(v),
\qquad
\Zeta_i^p=\operatorname{sg}\!\left[\hat{\Zeta}_{\mathrm{total}}\right].
\]
These quantities have the same forward value, so $\bar p_i(v)=p_i(v)/\Zeta_i^p$ sums to one over the support. During differentiation, $\nabla_\theta\Zeta_i^p=0$, while $\hat{\Zeta}_{\mathrm{total}}$ remains differentiable through the student probabilities. The teacher probabilities and selected token indices are fixed.

\paragraph{Gradient decomposition.}
Applying the product rule to the student weight and log-ratio in Eq.~\ref{eq:rkl-topk}, holding the teacher and detached denominator fixed and using $\nabla_\theta p_i(v)=p_i(v)\nabla_\theta\log p_i(v)$, gives
\begin{align}
\nabla_\theta\ell_i^{\mathrm{RKL}}
&=\sum_{v\in\mathcal V_K^{(i)}}\left[
\log\frac{p_i(v)}{q_i(v)}\,
\nabla_\theta\bar p_i(v)
+\bar p_i(v)
\nabla_\theta\log\frac{p_i(v)}{q_i(v)}
\right] \notag\\
&=\frac{1}{\Zeta_i^p}\sum_{v\in\mathcal V_K^{(i)}}
\left[
\log\frac{p_i(v)}{q_i(v)}\nabla_\theta p_i(v)
+p_i(v)\nabla_\theta\log p_i(v)
\right] \notag\\
&=\sum_{v\in\mathcal V_K^{(i)}}\frac{p_i(v)}{\Zeta_i^p}
\left(\log\frac{p_i(v)}{q_i(v)}+1\right)
\nabla_\theta\log p_i(v) \notag\\
&=\underbrace{\sum_{v\in\mathcal V_K^{(i)}}\bar p_i(v)
\log\frac{p_i(v)}{q_i(v)}\nabla_\theta\log p_i(v)}_{\mathbf g_i}
+\underbrace{\vphantom{\sum_{v\in\mathcal V_K^{(i)}}\bar p_i(v)\log\frac{p_i(v)}{q_i(v)}\nabla_\theta\log p_i(v)}
\nabla_\theta\log\hat{\Zeta}_{\mathrm{total}}}_{\mathbf h_i}.
\label{eq:wf-app-split}
\end{align}
Thus $\mathbf g_i$ comes from differentiating the student weights and retains the teacher-dependent log-ratios. The term $\mathbf h_i$ comes from differentiating the log-ratios themselves and is independent of the teacher. To see why it is a support-mass term, sum the probability derivatives over the retained tokens:
\begin{equation}
\begin{aligned}
\mathbf h_i
&=\sum_{v\in\mathcal V_K^{(i)}}\frac{p_i(v)}{\Zeta_i^p}\nabla_\theta\log p_i(v)
=\frac{1}{\Zeta_i^p}\sum_{v\in\mathcal V_K^{(i)}}\nabla_\theta p_i(v)\\
&=\frac{1}{\Zeta_i^p}\nabla_\theta\hat{\Zeta}_{\mathrm{total}}
=\frac{1}{\hat{\Zeta}_{\mathrm{total}}}\nabla_\theta\hat{\Zeta}_{\mathrm{total}}
=\nabla_\theta\log\hat{\Zeta}_{\mathrm{total}}.
\end{aligned}
\label{eq:wf-app-support-term}
\end{equation}
The simplification uses the equality of the forward values of $\Zeta_i^p$ and $\hat{\Zeta}_{\mathrm{total}}$; it does not differentiate the detached normalizer. Thus $-\mathbf h_i$ is a descent direction for the retained mass whenever $\mathbf h_i\ne0$. The full update also includes $\mathbf g_i$, which can counteract this decrease. Although $\bar p_i(v)$ sums to one in the forward pass, its gradients need not sum to zero. Over the full vocabulary, $\hat{\Zeta}_{\mathrm{total}}=1$ identically and $\mathbf h_i=0$. Truncation removes this cancellation, although $\mathbf h_i$ can still vanish where retained mass is locally stationary in $\theta$.

\subsection{Expected Gradient of the Sampled-Token Loss}
\label{app:wf-sampled-token}

We use the fixed behavior distribution $p_i^{\mathrm{old}}$ from Section~\ref{sec:objective} and assume it assigns positive probability wherever $p_i$ is positive.  Differentiating Eq.~\ref{eq:sampled-token-loss}, using its detached advantage and exact, unclipped importance ratio, and taking the expectation gives
\begin{align}
\mathbb E_{y_i\sim p_i^{\mathrm{old}}}\!\left[\nabla_\theta\ell_i^{\mathrm{ST}}\right]
&=\mathbb E_{y_i\sim p_i^{\mathrm{old}}}\!\left[
\frac{p_i(y_i)}{p_i^{\mathrm{old}}(y_i)}
\log\frac{p_i(y_i)}{q_i(y_i)}\nabla_\theta\log p_i(y_i)
\right] \notag\\
&=\sum_{v\in\mathcal V}p_i(v)\log\frac{p_i(v)}{q_i(v)}\nabla_\theta\log p_i(v) \notag\\
&=\nabla_\theta D_{\mathrm{KL}}(p_i\Vert q_i).
\label{eq:wf-st-expectation}
\end{align}
The importance ratio cancels the behavior probabilities in the expectation. The final equality uses $\sum_{v\in\mathcal V}\nabla_\theta p_i(v)=0$; at $p_i=q_i$, the ST gradient vanishes for every sampled action.
The expected ST parameter gradient therefore equals the full-vocabulary reverse-KL gradient, including when $p_i^{\mathrm{old}}\ne p_i$. Every token's own teacher score enters this expectation, including tokens outside the current top-$K$. Removing $\mathbf h_i$ from the truncated gradient alone leaves a restricted, renormalized sum and does not recover this expectation. These identities concern the fixed-prefix gradient and do not differentiate through the distribution of visited prefixes.

\subsection{Logit-Level Mechanism and Constructed Example}
\label{app:wf-constructed-example}

\paragraph{Differentiating the original loss.}
Let $z_i(j)$ be the student's logit for vocabulary token $j$ at the fixed prefix. Starting from Eq.~\ref{eq:rkl-topk}, we write the loss together with the full-vocabulary softmax:
\begin{equation}
\ell_i^{\mathrm{RKL}}
=\sum_{v\in\mathcal V_K^{(i)}}\frac{p_i(v)}{\Zeta_i^p}\log\frac{p_i(v)}{q_i(v)},
\qquad
p_i(v)=\frac{e^{z_i(v)}}{\sum_{v'\in\mathcal V}e^{z_i(v')}}.
\label{eq:wf-app-logit-loss}
\end{equation}
The selected support, teacher probabilities, and detached normalizer $\Zeta_i^p$ are held fixed during differentiation. The softmax denominator remains differentiable. Applying the quotient rule and then differentiating the log-ratio gives
\begin{align}
\frac{\partial p_i(v)}{\partial z_i(j)}
&=\frac{\mathds{1}[v=j]e^{z_i(v)}\sum_{v'\in\mathcal V}e^{z_i(v')}
-e^{z_i(v)}e^{z_i(j)}}{\left(\sum_{v'\in\mathcal V}e^{z_i(v')}\right)^2} \notag\\
&=p_i(v)\bigl(\mathds{1}[v=j]-p_i(j)\bigr), \notag\\
\frac{\partial}{\partial z_i(j)}\log\frac{p_i(v)}{q_i(v)}
&=\frac{1}{p_i(v)}\frac{\partial p_i(v)}{\partial z_i(j)}
=\mathds{1}[v=j]-p_i(j).
\label{eq:wf-app-softmax-derivatives}
\end{align}
Thus a logit outside the selected support still affects every retained probability. Applying the product rule directly to the loss in Eq.~\ref{eq:wf-app-logit-loss} yields
\begin{align}
\frac{\partial\ell_i^{\mathrm{RKL}}}{\partial z_i(j)}
&=\frac{1}{\Zeta_i^p}\sum_{v\in\mathcal V_K^{(i)}}\left[
\log\frac{p_i(v)}{q_i(v)}\frac{\partial p_i(v)}{\partial z_i(j)}
+p_i(v)\frac{\partial}{\partial z_i(j)}\log\frac{p_i(v)}{q_i(v)}
\right] \notag\\
&=\sum_{v\in\mathcal V_K^{(i)}}\frac{p_i(v)}{\Zeta_i^p}
\left(\log\frac{p_i(v)}{q_i(v)}+1\right)
\bigl(\mathds{1}[v=j]-p_i(j)\bigr) \notag\\
&=\mathds{1}[j\in\mathcal V_K^{(i)}]\frac{p_i(j)}{\Zeta_i^p}
\left(\log\frac{p_i(j)}{q_i(j)}+1\right) - p_i(j)\sum_{v\in\mathcal V_K^{(i)}}\frac{p_i(v)}{\Zeta_i^p}
\left(\log\frac{p_i(v)}{q_i(v)}+1\right).
\label{eq:wf-app-logit-gradient}
\end{align}
Both occurrences of $+1$ arise from differentiating the log-ratio. Using the forward values $\sum_{v\in\mathcal V_K^{(i)}}p_i(v)/\Zeta_i^p=1$ and Eq.~\ref{eq:wf-app-logit-loss}, the remaining weighted sum equals $\ell_i^{\mathrm{RKL}}+1$. The derivative therefore separates into two cases:
\begin{equation}
\frac{\partial\ell_i^{\mathrm{RKL}}}{\partial z_i(j)}
=\begin{cases}
\displaystyle
p_i(j)\left[\frac{\log[p_i(j)/q_i(j)]+1}{\Zeta_i^p}
-\bigl(\ell_i^{\mathrm{RKL}}+1\bigr)\right],
&j\in\mathcal V_K^{(i)},\\[8pt]
\displaystyle
-p_i(j)\bigl(\ell_i^{\mathrm{RKL}}+1\bigr),
&j\notin\mathcal V_K^{(i)}.
\end{cases}
\label{eq:wf-app-logit-cases}
\end{equation}
For a retained token, the derivative includes its own teacher log-ratio and the aggregate contribution from all retained tokens. For an omitted token, its own teacher probability $q_i(j)$ is absent: its derivative is determined by $p_i(j)$ and the retained aggregate. Under a logit descent step, omitted logits increase when $\ell_i^{\mathrm{RKL}}>-1$, stay unchanged at $-1$, and decrease below $-1$. This condition describes logit direction; support-mass transport additionally depends on the retained-token updates through the full softmax.

\paragraph{Updates used in the constructed example.}
For the RKL loss, we apply $z_i^+=z_i-\eta\nabla_{z_i}\ell_i^{\mathrm{RKL}}$ using the derivatives above and recompute $p_i^+=\operatorname{softmax}(z_i^+)$. For ST training, Eq.~\ref{eq:wf-st-expectation} and the same softmax derivatives give
\begin{equation}
\mathbb E\!\left[\frac{\partial\ell_i^{\mathrm{ST}}}{\partial z_i(j)}\right]
=p_i(j)\left[\log\frac{p_i(j)}{q_i(j)}-D_{\mathrm{KL}}(p_i\Vert q_i)\right].
\label{eq:wf-app-st-logit}
\end{equation}
We use this exact expected gradient for the ST updates. For the neural student, the chain rule gives
\[
\nabla_\theta\ell_i^{\mathrm{RKL}}
=\sum_{j\in\mathcal V}\frac{\partial\ell_i^{\mathrm{RKL}}}{\partial z_i(j)}\nabla_\theta z_i(j),
\]
and training aggregates these parameter gradients across supervised positions.

\begin{figure}[!htbp]
\centering
\includegraphics[width=\linewidth]{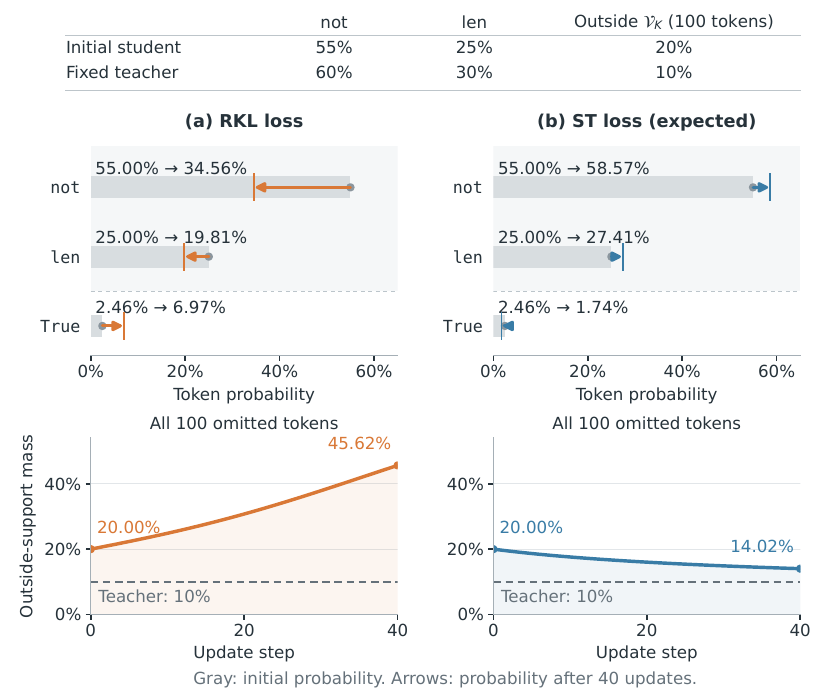}
\caption{\textbf{Constructed comparison of independent-logit updates.}
Upper panels show initial probabilities (gray) and final probabilities (arrow tips) of two retained tokens and one omitted token.
Lower panels track total outside-support mass; dashed lines mark the teacher.
ST updates use the exact expected gradient.}
\label{fig:wf-mechanism}
\end{figure}

Figure~\ref{fig:wf-mechanism} uses a 102-token vocabulary. Initially $p_i(v_1)=0.55$, $p_i(v_2)=0.25$, $q_i(v_1)=0.60$, and $q_i(v_2)=0.30$. For tail rank $k=1,\ldots,100$,
\begin{equation}
p_i(v_{k+2})=0.20\,\frac{k^{-0.8}}{\sum_{k'=1}^{100}(k')^{-0.8}},
\qquad
q_i(v_{k+2})=0.10\,\frac{k^{-1}}{\sum_{k'=1}^{100}(k')^{-1}}.
\end{equation}
Both trajectories take 40 independent-logit steps of size $0.25$. The same two tokens remain top-$2$ at every step. ST updates use the exact full-vocabulary expectation. After every step, probabilities and full-vocabulary KL are recomputed from the complete softmax; finite-difference checks hold $\Zeta_i^p$ fixed while recomputing the full softmax when verifying the truncated gradient.

Although the teacher assigns $90\%$ mass to the retained pair, compared with the student's initial $80\%$, RKL updates increase outside-support mass from $20.00\%$ to $45.62\%$. The surrogate falls from $-0.1168$ to $-0.5018$, while full-vocabulary reverse KL rises from $0.0537$ to $0.4297$. The expected ST update instead reduces outside-support mass to $14.02\%$ and full KL to $0.0145$.

This construction isolates loss reduction with worsening teacher alignment. Its initial support mass is chosen to make transport visible. Measuring the contribution to neural training additionally requires the logit Jacobian and updates at fixed diagnostic prefixes, since parameter gradients aggregate contributions across positions. The construction does not model the later concentration on repetitive sequences.

\end{document}

%% file: math_commands.tex
\usepackage{amsmath,amsfonts,bm}

\providecommand{\Zeta}{\mathrm{Z}}

\def\eqref#1{equation~\ref{#1}}
\def\1{\bm{1}}

\DeclareMathAlphabet{\mathsfit}{\encodingdefault}{\sfdefault}{m}{sl}
\SetMathAlphabet{\mathsfit}{bold}{\encodingdefault}{\sfdefault}{bx}{n}

%% file: iclr2027_conference.bib
@article{li2026rethinking,
  author       = {Yaxuan Li and
                  Yuxin Zuo and
                  Bingxiang He and
                  Jinqian Zhang and
                  Chaojun Xiao and
                  Cheng Qian and
                  Tianyu Yu and
                  Huan{-}ang Gao and
                  Wenkai Yang and
                  Zhiyuan Liu and
                  Ning Ding},
  title        = {Rethinking On-Policy Distillation of Large Language Models: Phenomenology,
                  Mechanism, and Recipe},
  journal      = {CoRR},
  volume       = {abs/2604.13016},
  year         = {2026},
  url          = {https://doi.org/10.48550/arXiv.2604.13016},
  doi          = {10.48550/ARXIV.2604.13016},
  eprinttype   = {arXiv},
  eprint       = {2604.13016},
  bibsource    = {dblp computer science bibliography, https://dblp.org}
}

@inproceedings{agarwal2024policy,
  author       = {Rishabh Agarwal and
                  Nino Vieillard and
                  Yongchao Zhou and
                  Piotr Stanczyk and
                  Sabela Ramos Garea and
                  Matthieu Geist and
                  Olivier Bachem},
  title        = {On-Policy Distillation of Language Models: Learning from Self-Generated
                  Mistakes},
  booktitle    = {The Twelfth International Conference on Learning Representations,
                  {ICLR} 2024, Vienna, Austria, May 7-11, 2024},
  publisher    = {OpenReview.net},
  year         = {2024},
  url          = {https://openreview.net/forum?id=3zKtaqxLhW},
  bibsource    = {dblp computer science bibliography, https://dblp.org}
}

@article{lu2025onpolicydistillation,
  author = {Kevin Lu and Thinking Machines Lab},
  title = {On-Policy Distillation},
  journal = {Thinking Machines Lab: Connectionism},
  year = {2025},
  note = {https://thinkingmachines.ai/blog/on-policy-distillation},
  doi = {10.64434/tml.20251026},
}

@article{yang2026learning,
  author       = {Wenkai Yang and
                  Weijie Liu and
                  Ruobing Xie and
                  Kai Yang and
                  Saiyong Yang and
                  Yankai Lin},
  title        = {Learning beyond Teacher: Generalized On-Policy Distillation with Reward
                  Extrapolation},
  journal      = {CoRR},
  volume       = {abs/2602.12125},
  year         = {2026},
  url          = {https://doi.org/10.48550/arXiv.2602.12125},
  doi          = {10.48550/ARXIV.2602.12125},
  eprinttype   = {arXiv},
  eprint       = {2602.12125},
  bibsource    = {dblp computer science bibliography, https://dblp.org}
}

@article{zhao2026self,
  author       = {Siyan Zhao and
                  Zhihui Xie and
                  Mengchen Liu and
                  Jing Huang and
                  Guan Pang and
                  Feiyu Chen and
                  Aditya Grover},
  title        = {Self-Distilled Reasoner: On-Policy Self-Distillation for Large Language
                  Models},
  journal      = {CoRR},
  volume       = {abs/2601.18734},
  year         = {2026},
  url          = {https://doi.org/10.48550/arXiv.2601.18734},
  doi          = {10.48550/ARXIV.2601.18734},
  eprinttype   = {arXiv},
  eprint       = {2601.18734},
  bibsource    = {dblp computer science bibliography, https://dblp.org}
}

@article{yang2026nemotron,
  author       = {Zhuolin Yang and
                  Zihan Liu and
                  Yang Chen and
                  Wenliang Dai and
                  Boxin Wang and
                  Sheng{-}Chieh Lin and
                  Chankyu Lee and
                  Yangyi Chen and
                  Dongfu Jiang and
                  Jiafan He and
                  Renjie Pi and
                  Grace Lam and
                  Nayeon Lee and
                  Alexander Bukharin and
                  Mohammad Shoeybi and
                  Bryan Catanzaro and
                  Wei Ping},
  title        = {Nemotron-Cascade 2: Post-Training LLMs with Cascade {RL} and Multi-Domain
                  On-Policy Distillation},
  journal      = {CoRR},
  volume       = {abs/2603.19220},
  year         = {2026},
  url          = {https://doi.org/10.48550/arXiv.2603.19220},
  doi          = {10.48550/ARXIV.2603.19220},
  eprinttype   = {arXiv},
  eprint       = {2603.19220},
  bibsource    = {dblp computer science bibliography, https://dblp.org}
}

@article{song2026survey,
  author       = {Mingyang Song and
                  Mao Zheng},
  title        = {A Survey of On-Policy Distillation for Large Language Models},
  journal      = {CoRR},
  volume       = {abs/2604.00626},
  year         = {2026},
  url          = {https://doi.org/10.48550/arXiv.2604.00626},
  doi          = {10.48550/ARXIV.2604.00626},
  eprinttype   = {arXiv},
  eprint       = {2604.00626},
  bibsource    = {dblp computer science bibliography, https://dblp.org}
}

@article{wu2026lightning,
  author       = {Yecheng Wu and
                  Song Han and
                  Hai Cai},
  title        = {Lightning {OPD:} Efficient Post-Training for Large Reasoning Models
                  with Offline On-Policy Distillation},
  journal      = {CoRR},
  volume       = {abs/2604.13010},
  year         = {2026},
  url          = {https://doi.org/10.48550/arXiv.2604.13010},
  doi          = {10.48550/ARXIV.2604.13010},
  eprinttype   = {arXiv},
  eprint       = {2604.13010},
  bibsource    = {dblp computer science bibliography, https://dblp.org}
}

@article{gou2021knowledge,
  author       = {Jianping Gou and
                  Baosheng Yu and
                  Stephen J. Maybank and
                  Dacheng Tao},
  title        = {Knowledge Distillation: {A} Survey},
  journal      = {Int. J. Comput. Vis.},
  volume       = {129},
  number       = {6},
  pages        = {1789--1819},
  year         = {2021},
  url          = {https://doi.org/10.1007/s11263-021-01453-z},
  doi          = {10.1007/S11263-021-01453-Z},
  bibsource    = {dblp computer science bibliography, https://dblp.org}
}

@article{hinton2015distilling,
  author       = {Geoffrey E. Hinton and
                  Oriol Vinyals and
                  Jeffrey Dean},
  title        = {Distilling the Knowledge in a Neural Network},
  journal      = {CoRR},
  volume       = {abs/1503.02531},
  year         = {2015},
  url          = {http://arxiv.org/abs/1503.02531},
  eprinttype   = {arXiv},
  eprint       = {1503.02531},
  bibsource    = {dblp computer science bibliography, https://dblp.org}
}

@inproceedings{ye2025analyzing,
  author       = {Junjie Ye and
                  Yuming Yang and
                  Yang Nan and
                  Shuo Li and
                  Qi Zhang and
                  Tao Gui and
                  Xuanjing Huang and
                  Peng Wang and
                  Zhongchao Shi and
                  Jianping Fan},
  editor       = {Christos Christodoulopoulos and
                  Tanmoy Chakraborty and
                  Carolyn Rose and
                  Violet Peng},
  title        = {Analyzing the Effects of Supervised Fine-Tuning on Model Knowledge
                  from Token and Parameter Levels},
  booktitle    = {Proceedings of the 2025 Conference on Empirical Methods in Natural
                  Language Processing, {EMNLP} 2025, Suzhou, China, November 4-9, 2025},
  pages        = {471--513},
  publisher    = {Association for Computational Linguistics},
  year         = {2025},
  url          = {https://doi.org/10.18653/v1/2025.emnlp-main.25},
  doi          = {10.18653/V1/2025.EMNLP-MAIN.25},
  bibsource    = {dblp computer science bibliography, https://dblp.org}
}

@article{team2026kimi,
  author       = {Kimi Team},
  title        = {Kimi {K3:} Open Frontier Intelligence},
  journal      = {CoRR},
  volume       = {abs/2607.24653},
  year         = {2026},
  url          = {https://doi.org/10.48550/arXiv.2607.24653},
  doi          = {10.48550/ARXIV.2607.24653},
  eprinttype   = {arXiv},
  eprint       = {2607.24653},
  bibsource    = {dblp computer science bibliography, https://dblp.org}
}

@article{yang2025qwen3,
  author       = {Qwen Team},
  title        = {Qwen3 Technical Report},
  journal      = {CoRR},
  volume       = {abs/2505.09388},
  year         = {2025},
  url          = {https://doi.org/10.48550/arXiv.2505.09388},
  doi          = {10.48550/ARXIV.2505.09388},
  eprinttype   = {arXiv},
  eprint       = {2505.09388},
  bibsource    = {dblp computer science bibliography, https://dblp.org}
}

@article{xiao2026mimo,
  author       = {LLM{-}Core Xiaomi},
  title        = {MiMo-V2-Flash Technical Report},
  journal      = {CoRR},
  volume       = {abs/2601.02780},
  year         = {2026},
  url          = {https://doi.org/10.48550/arXiv.2601.02780},
  doi          = {10.48550/ARXIV.2601.02780},
  eprinttype   = {arXiv},
  eprint       = {2601.02780},
  bibsource    = {dblp computer science bibliography, https://dblp.org}
}

@article{ma2026mopd,
  author       = {Wenhan Ma and
                  Jianyu Wei and
                  Liang Zhao and
                  Hailin Zhang and
                  Bangjun Xiao and
                  Lei Li and
                  Qibin Yang and
                  Bofei Gao and
                  Yudong Wang and
                  Rang Li and
                  Jinhao Dong and
                  Zhifang Sui and
                  Fuli Luo},
  title        = {{MOPD:} Multi-Teacher On-Policy Distillation for Capability Integration
                  in {LLM} Post-Training},
  journal      = {CoRR},
  volume       = {abs/2606.30406},
  year         = {2026},
  url          = {https://doi.org/10.48550/arXiv.2606.30406},
  doi          = {10.48550/ARXIV.2606.30406},
  eprinttype   = {arXiv},
  eprint       = {2606.30406},
  bibsource    = {dblp computer science bibliography, https://dblp.org}
}

@article{zeng2026glm,
  author       = {GLM},
  title        = {{GLM-5:} from Vibe Coding to Agentic Engineering},
  journal      = {CoRR},
  volume       = {abs/2602.15763},
  year         = {2026},
  url          = {https://doi.org/10.48550/arXiv.2602.15763},
  doi          = {10.48550/ARXIV.2602.15763},
  eprinttype   = {arXiv},
  eprint       = {2602.15763},
  bibsource    = {dblp computer science bibliography, https://dblp.org}
}

@article{xing2026trust,
  author       = {Xingrun Xing and
                  Haoqing Wang and
                  Boyan Gao and
                  Ziheng Li and
                  Yehui Tang},
  title        = {Trust Region On-Policy Distillation},
  journal      = {CoRR},
  volume       = {abs/2606.01249},
  year         = {2026},
  url          = {https://doi.org/10.48550/arXiv.2606.01249},
  doi          = {10.48550/ARXIV.2606.01249},
  eprinttype   = {arXiv},
  eprint       = {2606.01249},
  bibsource    = {dblp computer science bibliography, https://dblp.org}
}

@article{yang2026self,
  author       = {Chenxu Yang and
                  Chuanyu Qin and
                  Qingyi Si and
                  Minghui Chen and
                  Naibin Gu and
                  Dingyu Yao and
                  Zheng Lin and
                  Weiping Wang and
                  Jiaqi Wang and
                  Nan Duan},
  title        = {Self-Distilled {RLVR}},
  journal      = {CoRR},
  volume       = {abs/2604.03128},
  year         = {2026},
  url          = {https://doi.org/10.48550/arXiv.2604.03128},
  doi          = {10.48550/ARXIV.2604.03128},
  eprinttype   = {arXiv},
  eprint       = {2604.03128},
  bibsource    = {dblp computer science bibliography, https://dblp.org}
}

@article{hou2026uni,
  author       = {Wenjin Hou and
                  Shangpin Peng and
                  Weinong Wang and
                  Zheng Ruan and
                  Yue Zhang and
                  Zhenglin Zhou and
                  Mingqi Gao and
                  Yifei Chen and
                  Kaiqi Wang and
                  Hongming Yang and
                  Chengquan Zhang and
                  Zhuotao Tian and
                  Han Hu and
                  Yi Yang and
                  Fei Wu and
                  Hehe Fan},
  title        = {Uni-OPD: Unifying On-Policy Distillation with a Dual-Perspective Recipe},
  journal      = {CoRR},
  volume       = {abs/2605.03677},
  year         = {2026},
  url          = {https://doi.org/10.48550/arXiv.2605.03677},
  doi          = {10.48550/ARXIV.2605.03677},
  eprinttype   = {arXiv},
  eprint       = {2605.03677},
  bibsource    = {dblp computer science bibliography, https://dblp.org}
}

@article{zhou2026sage,
  author       = {Yuhang Zhou and
                  Lizhu Zhang and
                  Yifan Wu and
                  Mingyi Wang and
                  Bo Peng and
                  Jiayi Liu and
                  Xiangjun Fan and
                  Zhuokai Zhao},
  title        = {{SAGE-OPD:} Selective Agent-Guided Intervention for Multi-Turn On-Policy
                  Distillation},
  journal      = {CoRR},
  volume       = {abs/2606.19659},
  year         = {2026},
  url          = {https://doi.org/10.48550/arXiv.2606.19659},
  doi          = {10.48550/ARXIV.2606.19659},
  eprinttype   = {arXiv},
  eprint       = {2606.19659},
  bibsource    = {dblp computer science bibliography, https://dblp.org}
}

@article{wang2026tcod,
  author       = {Jiaqi Wang and
                  Wenhao Zhang and
                  Weijie Shi and
                  Yaliang Li and
                  James Cheng},
  title        = {{TCOD:} Exploring Temporal Curriculum in On-Policy Distillation for
                  Multi-turn Autonomous Agents},
  journal      = {CoRR},
  volume       = {abs/2604.24005},
  year         = {2026},
  url          = {https://doi.org/10.48550/arXiv.2604.24005},
  doi          = {10.48550/ARXIV.2604.24005},
  eprinttype   = {arXiv},
  eprint       = {2604.24005},
  bibsource    = {dblp computer science bibliography, https://dblp.org}
}

@article{wang2026mad,
  author       = {Jianze Wang and
                  Ying Liu and
                  Jinlong Chen and
                  Xuchun Hu and
                  Qilong Zhang and
                  Yu Cao and
                  Jun Wang and
                  Hua Yang and
                  Yong Xie and
                  Qianglong Chen},
  title        = {{MAD-OPD:} Breaking the Ceiling in On-Policy Distillation via Multi-Agent
                  Debate},
  journal      = {CoRR},
  volume       = {abs/2605.01347},
  year         = {2026},
  url          = {https://doi.org/10.48550/arXiv.2605.01347},
  doi          = {10.48550/ARXIV.2605.01347},
  eprinttype   = {arXiv},
  eprint       = {2605.01347},
  bibsource    = {dblp computer science bibliography, https://dblp.org}
}

@inproceedings{gu2023minillm,
  author       = {Yuxian Gu and
                  Li Dong and
                  Furu Wei and
                  Minlie Huang},
  title        = {MiniLLM: Knowledge Distillation of Large Language Models},
  booktitle    = {The Twelfth International Conference on Learning Representations,
                  {ICLR} 2024, Vienna, Austria, May 7-11, 2024},
  publisher    = {OpenReview.net},
  year         = {2024},
  url          = {https://openreview.net/forum?id=5h0qf7IBZZ},
  bibsource    = {dblp computer science bibliography, https://dblp.org}
}

@inproceedings{zan2026multi,
  author       = {Daoguang Zan and
                  Zhirong Huang and
                  Wei Liu and
                  Hanwu Chen and
                  Shulin Xin and
                  Linhao Zhang and
                  Qi Liu and
                  Aoyan Li and
                  Lu Chen and
                  Xiaojian Zhong and
                  Siyao Liu and
                  Yongsheng Xiao and
                  Liangqiang Chen and
                  Yuyu Zhang and
                  Jing Su and
                  Tianyu Liu and
                  Rui Long and
                  Ming Ding and
                  Liang Xiang},
  editor       = {Danielle Belgrave and
                  Cheng Zhang and
                  Laura N. Montoya and
                  Hsuan{-}Tien Lin and
                  Razvan Pascanu and
                  Piotr Koniusz and
                  Marzyeh Ghassemi and
                  Nancy Chen and
                  Iv{\'{a}}n Vladimir Meza Ru{\'{\i}}z and
                  Arturo Loaiza{-}Bonilla},
  title        = {Multi-SWE-bench: {A} Multilingual Benchmark for Issue Resolving},
  booktitle    = {Advances in Neural Information Processing Systems 38: Annual Conference
                  on Neural Information Processing Systems 2025, NeurIPS 2025, San Diego,
                  CA, USA, December 2-7, 2025 / Mexico City, Mexico, November 30 - December
                  5, 2025},
  year         = {2025},
  url          = {http://papers.nips.cc/paper\_files/paper/2025/hash/5afa9cb1e917b898ad418216dc726fbd-Abstract-Datasets\_and\_Benchmarks\_Track.html},
  bibsource    = {dblp computer science bibliography, https://dblp.org}
}

@inproceedings{jimenez2024swe,
  author       = {Carlos E. Jimenez and
                  John Yang and
                  Alexander Wettig and
                  Shunyu Yao and
                  Kexin Pei and
                  Ofir Press and
                  Karthik R. Narasimhan},
  title        = {SWE-bench: Can Language Models Resolve Real-world Github Issues?},
  booktitle    = {The Twelfth International Conference on Learning Representations,
                  {ICLR} 2024, Vienna, Austria, May 7-11, 2024},
  publisher    = {OpenReview.net},
  year         = {2024},
  url          = {https://openreview.net/forum?id=VTF8yNQM66},
  bibsource    = {dblp computer science bibliography, https://dblp.org}
}

@article{deng2025swe,
  author       = {Xiang Deng and
                  Jeff Da and
                  Edwin Pan and
                  Yannis Yiming He and
                  Charles Ide and
                  Kanak Garg and
                  Niklas Lauffer and
                  Andrew Park and
                  Nitin Pasari and
                  Chetan Rane and
                  Karmini Sampath and
                  Maya Krishnan and
                  Srivatsa Kundurthy and
                  Sean Hendryx and
                  Zifan Wang and
                  Chen Bo Calvin Zhang and
                  Noah Jacobson and
                  Bing Liu and
                  Brad Kenstler},
  title        = {SWE-Bench Pro: Can {AI} Agents Solve Long-Horizon Software Engineering
                  Tasks?},
  journal      = {CoRR},
  volume       = {abs/2509.16941},
  year         = {2025},
  url          = {https://doi.org/10.48550/arXiv.2509.16941},
  doi          = {10.48550/ARXIV.2509.16941},
  eprinttype   = {arXiv},
  eprint       = {2509.16941},
  bibsource    = {dblp computer science bibliography, https://dblp.org}
}

@article{shao2024deepseekmath,
  author       = {Zhihong Shao and
                  Peiyi Wang and
                  Qihao Zhu and
                  Runxin Xu and
                  Junxiao Song and
                  Mingchuan Zhang and
                  Y. K. Li and
                  Y. Wu and
                  Daya Guo},
  title        = {DeepSeekMath: Pushing the Limits of Mathematical Reasoning in Open
                  Language Models},
  journal      = {CoRR},
  volume       = {abs/2402.03300},
  year         = {2024},
  url          = {https://doi.org/10.48550/arXiv.2402.03300},
  doi          = {10.48550/ARXIV.2402.03300},
  eprinttype   = {arXiv},
  eprint       = {2402.03300},
  bibsource    = {dblp computer science bibliography, https://dblp.org}
}

@article{wang2026cross,
  author       = {Hao Wang and
                  Kun Yuan and
                  Wenlin Zhong and
                  Minglei Zhang and
                  Han Xiao and
                  Ming Sun and
                  Honggang Qi},
  title        = {Cross-Tokenizer On-Policy Distillation via Byte-Prefix Marginalization},
  journal      = {CoRR},
  volume       = {abs/2607.22334},
  year         = {2026},
  url          = {https://doi.org/10.48550/arXiv.2607.22334},
  doi          = {10.48550/ARXIV.2607.22334},
  eprinttype   = {arXiv},
  eprint       = {2607.22334},
  bibsource    = {dblp computer science bibliography, https://dblp.org}
}

@article{sun2026simct,
  author       = {Jie Sun and
                  Mao Zheng and
                  Mingyang Song and
                  Qiyong Zhong and
                  Yilin Cheng and
                  Bichuan Feng and
                  Pengfei Liu and
                  Junfeng Fang and
                  Xiang Wang},
  title        = {SimCT: Recovering Lost Supervision for Cross-Tokenizer On-Policy Distillation},
  journal      = {CoRR},
  volume       = {abs/2605.07711},
  year         = {2026},
  url          = {https://doi.org/10.48550/arXiv.2605.07711},
  doi          = {10.48550/ARXIV.2605.07711},
  eprinttype   = {arXiv},
  eprint       = {2605.07711},
  bibsource    = {dblp computer science bibliography, https://dblp.org}
}

@misc{pati2025,
  title={Unlocking On-Policy Distillation for Any Model Family},
  author={Carlos Miguel Patiño and Kashif Rasul and Quentin Gallouédec and Ben Burtenshaw and Sergio Paniego and Vaibhav Srivastav and Thibaud Frere and Ed Beeching and Lewis Tunstall and Leandro von Werra and Thomas Wolf},
  year={2025},
  howpublished={\url{https://huggingface.co/spaces/HuggingFaceH4/on-policy-distillation}},
  
}

@article{niu2026breaking,
  author       = {Yifan Niu and
                  Han Xiao and
                  Dongyi Liu and
                  Zelong Wang and
                  Dihong Gong and
                  Yasheng Wang and
                  Jia Li},
  title        = {Breaking the Tokenizer Barrier: On-Policy Distillation across Model
                  Families},
  journal      = {CoRR},
  volume       = {abs/2606.09456},
  year         = {2026},
  url          = {https://doi.org/10.48550/arXiv.2606.09456},
  doi          = {10.48550/ARXIV.2606.09456},
  eprinttype   = {arXiv},
  eprint       = {2606.09456},
  bibsource    = {dblp computer science bibliography, https://dblp.org}
}

@misc{uniagent_github,
  author       = {Yuyang Ding and Bo Wen and Xubo Cao and Zhiqiang Zhai and Guangming Sheng and Xibin Wu and Juntao Li and Min Zhang and Uni-Agent Contributors},
  title        = {Uni-Agent: Build, Run, and Train Agents at Scale},
  year         = {2026},
  howpublished = {\url{https://github.com/verl-project/uni-agent}},
  note         = {GitHub repository. Supervisor: Xibin Wu and Juntao Li},
  urldate      = {2026-03-27}
}

@misc{slime_github,
  author       = {Zilin Zhu and Chengxing Xie and Xin Lv and slime Contributors},
  title        = {slime: An LLM post-training framework for RL Scaling},
  year         = {2025},
  howpublished = {\url{https://github.com/THUDM/slime}},
  note         = {GitHub repository. Corresponding author: Xin Lv},
  urldate      = {2025-06-19}
}

@inproceedings{zheng2024sglang,
  author       = {Lianmin Zheng and
                  Liangsheng Yin and
                  Zhiqiang Xie and
                  Chuyue Sun and
                  Jeff Huang and
                  Cody Hao Yu and
                  Shiyi Cao and
                  Christos Kozyrakis and
                  Ion Stoica and
                  Joseph E. Gonzalez and
                  Clark W. Barrett and
                  Ying Sheng},
  editor       = {Amir Globersons and
                  Lester Mackey and
                  Danielle Belgrave and
                  Angela Fan and
                  Ulrich Paquet and
                  Jakub M. Tomczak and
                  Cheng Zhang},
  title        = {SGLang: Efficient Execution of Structured Language Model Programs},
  booktitle    = {Advances in Neural Information Processing Systems 37: Annual Conference
                  on Neural Information Processing Systems 2024, NeurIPS 2024, Vancouver,
                  BC, Canada, December 10 - 15, 2024},
  year         = {2024},
  url          = {http://papers.nips.cc/paper\_files/paper/2024/hash/724be4472168f31ba1c9ac630f15dec8-Abstract-Conference.html},
  bibsource    = {dblp computer science bibliography, https://dblp.org}
}

@article{jain2025r2e,
  author       = {Naman Jain and
                  Jaskirat Singh and
                  Manish Shetty and
                  Liang Zheng and
                  Koushik Sen and
                  Ion Stoica},
  title        = {R2E-Gym: Procedural Environments and Hybrid Verifiers for Scaling
                  Open-Weights {SWE} Agents},
  journal      = {CoRR},
  volume       = {abs/2504.07164},
  year         = {2025},
  url          = {https://doi.org/10.48550/arXiv.2504.07164},
  doi          = {10.48550/ARXIV.2504.07164},
  eprinttype   = {arXiv},
  eprint       = {2504.07164},
  bibsource    = {dblp computer science bibliography, https://dblp.org}
}

@article{zhao2026immersiongithubuniversescaling,
  author       = {Jiale Zhao and
                  Guoxin Chen and
                  Fanzhe Meng and
                  Minghao Li and
                  Jie Chen and
                  Hui Xu and
                  Yongshuai Sun and
                  Xin Zhao and
                  Ruihua Song and
                  Yuan Zhang and
                  Peng Wang and
                  Cheng Chen and
                  Jirong Wen and
                  Kai Jia},
  title        = {Immersion in the GitHub Universe: Scaling Coding Agents to Mastery},
  journal      = {CoRR},
  volume       = {abs/2602.09892},
  year         = {2026},
  url          = {https://doi.org/10.48550/arXiv.2602.09892},
  doi          = {10.48550/ARXIV.2602.09892},
  eprinttype   = {arXiv},
  eprint       = {2602.09892},
  bibsource    = {dblp computer science bibliography, https://dblp.org}
}

@article{guo2026menvagent,
  author       = {Chuanzhe Guo and
                  Jingjing Wu and
                  Sijun He and
                  Yang Chen and
                  Zhaoqi Kuang and
                  Shilong Fan and
                  Bingjin Chen and
                  Siqi Bao and
                  Jing Liu and
                  Hua Wu and
                  Qingfu Zhu and
                  Wanxiang Che and
                  Haifeng Wang},
  title        = {MEnvAgent: Scalable Polyglot Environment Construction for Verifiable
                  Software Engineering},
  journal      = {CoRR},
  volume       = {abs/2601.22859},
  year         = {2026},
  url          = {https://doi.org/10.48550/arXiv.2601.22859},
  doi          = {10.48550/ARXIV.2601.22859},
  eprinttype   = {arXiv},
  eprint       = {2601.22859},
  bibsource    = {dblp computer science bibliography, https://dblp.org}
}

@inproceedings{kim2016sequence,
  author       = {Yoon Kim and
                  Alexander M. Rush},
  editor       = {Jian Su and
                  Xavier Carreras and
                  Kevin Duh},
  title        = {Sequence-Level Knowledge Distillation},
  booktitle    = {Proceedings of the 2016 Conference on Empirical Methods in Natural
                  Language Processing, {EMNLP} 2016, Austin, Texas, USA, November 1-4,
                  2016},
  pages        = {1317--1327},
  publisher    = {The Association for Computational Linguistics},
  year         = {2016},
  url          = {https://doi.org/10.18653/v1/d16-1139},
  doi          = {10.18653/V1/D16-1139},
  bibsource    = {dblp computer science bibliography, https://dblp.org}
}

@inproceedings{rusu2016policy,
  author       = {Andrei A. Rusu and
                  Sergio Gomez Colmenarejo and
                  {\c{C}}aglar G{\"{u}}l{\c{c}}ehre and
                  Guillaume Desjardins and
                  James Kirkpatrick and
                  Razvan Pascanu and
                  Volodymyr Mnih and
                  Koray Kavukcuoglu and
                  Raia Hadsell},
  editor       = {Yoshua Bengio and
                  Yann LeCun},
  title        = {Policy Distillation},
  booktitle    = {4th International Conference on Learning Representations, {ICLR} 2016,
                  San Juan, Puerto Rico, May 2-4, 2016, Conference Track Proceedings},
  year         = {2016},
  url          = {http://arxiv.org/abs/1511.06295},
  bibsource    = {dblp computer science bibliography, https://dblp.org}
}

@inproceedings{lin2020autoregressive,
  author       = {Alexander Lin and
                  Jeremy Wohlwend and
                  Howard Chen and
                  Tao Lei},
  editor       = {Bonnie Webber and
                  Trevor Cohn and
                  Yulan He and
                  Yang Liu},
  title        = {Autoregressive Knowledge Distillation through Imitation Learning},
  booktitle    = {Proceedings of the 2020 Conference on Empirical Methods in Natural
                  Language Processing, {EMNLP} 2020, Online, November 16-20, 2020},
  pages        = {6121--6133},
  publisher    = {Association for Computational Linguistics},
  year         = {2020},
  url          = {https://doi.org/10.18653/v1/2020.emnlp-main.494},
  doi          = {10.18653/V1/2020.EMNLP-MAIN.494},
  bibsource    = {dblp computer science bibliography, https://dblp.org}
}

@inproceedings{agarwal2024gkd,
  author       = {Rishabh Agarwal and
                  Nino Vieillard and
                  Yongchao Zhou and
                  Piotr Stanczyk and
                  Sabela Ramos Garea and
                  Matthieu Geist and
                  Olivier Bachem},
  title        = {On-Policy Distillation of Language Models: Learning from Self-Generated
                  Mistakes},
  booktitle    = {The Twelfth International Conference on Learning Representations,
                  {ICLR} 2024, Vienna, Austria, May 7-11, 2024},
  publisher    = {OpenReview.net},
  year         = {2024},
  url          = {https://openreview.net/forum?id=3zKtaqxLhW},
  bibsource    = {dblp computer science bibliography, https://dblp.org}
}

@article{deepseekr1,
  author       = {Daya Guo and
                  Dejian Yang and
                  Haowei Zhang and
                  Junxiao Song and
                  Peiyi Wang and
                  Qihao Zhu and
                  Runxin Xu and
                  Ruoyu Zhang and
                  Shirong Ma and
                  Xiao Bi and
                  Xiaokang Zhang and
                  Xingkai Yu and
                  Yu Wu and
                  Z. F. Wu and
                  Zhibin Gou and
                  Zhihong Shao and
                  Zhuoshu Li and
                  Ziyi Gao and
                  Aixin Liu and
                  Bing Xue and
                  Bingxuan Wang and
                  Bochao Wu and
                  Bei Feng and
                  Chengda Lu and
                  Chenggang Zhao and
                  Chengqi Deng and
                  Chong Ruan and
                  Damai Dai and
                  Deli Chen and
                  Dongjie Ji and
                  Erhang Li and
                  Fangyun Lin and
                  Fucong Dai and
                  Fuli Luo and
                  Guangbo Hao and
                  Guanting Chen and
                  Guowei Li and
                  Hao Zhang and
                  Hanwei Xu and
                  Honghui Ding and
                  Huazuo Gao and
                  Hui Qu and
                  Hui Li and
                  Jianzhong Guo and
                  Jiashi Li and
                  Jingchang Chen and
                  Jingyang Yuan and
                  Jinhao Tu and
                  Junjie Qiu and
                  Junlong Li and
                  J. L. Cai and
                  Jiaqi Ni and
                  Jian Liang and
                  Jin Chen and
                  Kai Dong and
                  Kai Hu and
                  Kaichao You and
                  Kaige Gao and
                  Kang Guan and
                  Kexin Huang and
                  Kuai Yu and
                  Lean Wang and
                  Lecong Zhang and
                  Liang Zhao and
                  Litong Wang and
                  Liyue Zhang and
                  Lei Xu and
                  Leyi Xia and
                  Mingchuan Zhang and
                  Minghua Zhang and
                  Minghui Tang and
                  Mingxu Zhou and
                  Meng Li and
                  Miaojun Wang and
                  Mingming Li and
                  Ning Tian and
                  Panpan Huang and
                  Peng Zhang and
                  Qiancheng Wang and
                  Qinyu Chen and
                  Qiushi Du and
                  Ruiqi Ge and
                  Ruisong Zhang and
                  Ruizhe Pan and
                  Runji Wang and
                  R. J. Chen and
                  R. L. Jin and
                  Ruyi Chen and
                  Shanghao Lu and
                  Shangyan Zhou and
                  Shanhuang Chen and
                  Shengfeng Ye and
                  Shiyu Wang and
                  Shuiping Yu and
                  Shunfeng Zhou and
                  Shuting Pan and
                  S. S. Li and
                  Shuang Zhou and
                  Shaoqing Wu and
                  Tao Yun and
                  Tian Pei and
                  Tianyu Sun and
                  Tao Wang and
                  Wangding Zeng and
                  Wen Liu and
                  Wenfeng Liang and
                  Wenjun Gao and
                  Wenqin Yu and
                  Wentao Zhang and
                  W. L. Xiao and
                  Wei An and
                  Xiaodong Liu and
                  Xiaohan Wang and
                  Xiaokang Chen and
                  Xiaotao Nie and
                  Xin Cheng and
                  Xin Liu and
                  Xin Xie and
                  Xingchao Liu and
                  Xinyu Yang and
                  Xinyuan Li and
                  Xuecheng Su and
                  Xuheng Lin and
                  X. Q. Li and
                  Xiangyue Jin and
                  Xiaojin Shen and
                  Xiaosha Chen and
                  Xiaowen Sun and
                  Xiaoxiang Wang and
                  Xinnan Song and
                  Xinyi Zhou and
                  Xianzu Wang and
                  Xinxia Shan and
                  Y. K. Li and
                  Y. Q. Wang and
                  Y. X. Wei and
                  Yang Zhang and
                  Yanhong Xu and
                  Yao Li and
                  Yao Zhao and
                  Yaofeng Sun and
                  Yaohui Wang and
                  Yi Yu and
                  Yichao Zhang and
                  Yifan Shi and
                  Yiliang Xiong and
                  Ying He and
                  Yishi Piao and
                  Yisong Wang and
                  Yixuan Tan and
                  Yiyang Ma and
                  Yiyuan Liu and
                  Yongqiang Guo and
                  Yuan Ou and
                  Yuduan Wang and
                  Yue Gong and
                  Yuheng Zou and
                  Yujia He and
                  Yunfan Xiong and
                  Yuxiang Luo and
                  Yuxiang You and
                  Yuxuan Liu and
                  Yuyang Zhou and
                  Y. X. Zhu and
                  Yanping Huang and
                  Yaohui Li and
                  Yi Zheng and
                  Yuchen Zhu and
                  Yunxian Ma and
                  Ying Tang and
                  Yukun Zha and
                  Yuting Yan and
                  Z. Z. Ren and
                  Zehui Ren and
                  Zhangli Sha and
                  Zhe Fu and
                  Zhean Xu and
                  Zhenda Xie and
                  Zhengyan Zhang and
                  Zhewen Hao and
                  Zhicheng Ma and
                  Zhigang Yan and
                  Zhiyu Wu and
                  Zihui Gu and
                  Zijia Zhu and
                  Zijun Liu and
                  Zilin Li and
                  Ziwei Xie and
                  Ziyang Song and
                  Zizheng Pan and
                  Zhen Huang and
                  Zhipeng Xu and
                  Zhongyu Zhang and
                  Zhen Zhang},
  title        = {DeepSeek-R1 incentivizes reasoning in LLMs through reinforcement learning},
  journal      = {Nat.},
  volume       = {645},
  number       = {8081},
  pages        = {633--638},
  year         = {2025},
  url          = {https://doi.org/10.1038/s41586-025-09422-z},
  doi          = {10.1038/S41586-025-09422-Z},
  bibsource    = {dblp computer science bibliography, https://dblp.org}
}

@inproceedings{ko2024distillm,
  author       = {Jongwoo Ko and
                  Sungnyun Kim and
                  Tianyi Chen and
                  Se{-}Young Yun},
  editor       = {Ruslan Salakhutdinov and
                  Zico Kolter and
                  Katherine A. Heller and
                  Adrian Weller and
                  Nuria Oliver and
                  Jonathan Scarlett and
                  Felix Berkenkamp},
  title        = {DistiLLM: Towards Streamlined Distillation for Large Language Models},
  booktitle    = {Forty-first International Conference on Machine Learning, {ICML} 2024,
                  Vienna, Austria, July 21-27, 2024},
  series       = {Proceedings of Machine Learning Research},
  volume       = {235},
  pages        = {24872--24895},
  publisher    = {{PMLR} / OpenReview.net},
  year         = {2024},
  url          = {https://proceedings.mlr.press/v235/ko24c.html},
  bibsource    = {dblp computer science bibliography, https://dblp.org}
}

@article{he2026simpleopdsimpletokenizeragnosticonpolicy,
  author       = {Haonan He and
                  Haodi Lei and
                  Yun Luo and
                  Haoran Zhang and
                  Shunkai Zhang and
                  Yizhuo Li and
                  Shengji Tang and
                  Zhilin Wang and
                  Runzhe Zhan and
                  Lei Bai and
                  Ganqu Cui and
                  Fangchen Yu and
                  Yafu Li and
                  Peng Ye and
                  Ning Ding and
                  Yu Cheng},
  title        = {SimpleOPD: Simple Tokenizer-Agnostic On-Policy Distillation for Long-Context
                  Reasoning},
  journal      = {CoRR},
  volume       = {abs/2608.14277},
  year         = {2026},
  url          = {https://doi.org/10.48550/arXiv.2608.14277},
  doi          = {10.48550/ARXIV.2608.14277},
  eprinttype   = {arXiv},
  eprint       = {2608.14277},
  bibsource    = {dblp computer science bibliography, https://dblp.org}
}

@article{DBLP:journals/corr/abs-2306-13649,
  author       = {Rishabh Agarwal and
                  Nino Vieillard and
                  Piotr Stanczyk and
                  Sabela Ramos and
                  Matthieu Geist and
                  Olivier Bachem},
  title        = {{GKD:} Generalized Knowledge Distillation for Auto-regressive Sequence
                  Models},
  journal      = {CoRR},
  volume       = {abs/2306.13649},
  year         = {2023},
  url          = {https://doi.org/10.48550/arXiv.2306.13649},
  doi          = {10.48550/ARXIV.2306.13649},
  eprinttype   = {arXiv},
  eprint       = {2306.13649},
  bibsource    = {dblp computer science bibliography, https://dblp.org}
}

@inproceedings{DBLP:journals/jmlr/RossGB11,
  author       = {St{\'{e}}phane Ross and
                  Geoffrey J. Gordon and
                  Drew Bagnell},
  editor       = {Geoffrey J. Gordon and
                  David B. Dunson and
                  Miroslav Dud{\'{\i}}k},
  title        = {A Reduction of Imitation Learning and Structured Prediction to No-Regret
                  Online Learning},
  booktitle    = {Proceedings of the Fourteenth International Conference on Artificial
                  Intelligence and Statistics, {AISTATS} 2011, Fort Lauderdale, USA,
                  April 11-13, 2011},
  series       = {{JMLR} Proceedings},
  volume       = {15},
  pages        = {627--635},
  publisher    = {JMLR.org},
  year         = {2011},
  url          = {http://proceedings.mlr.press/v15/ross11a/ross11a.pdf},
  bibsource    = {dblp computer science bibliography, https://dblp.org}
}

@misc{2025_unlocking_on_policy_distillation_for_any_model_family,
  title={Unlocking On-Policy Distillation for Any Model Family},
  author={Carlos Miguel Patiño and Kashif Rasul and Quentin Gallouédec and Ben Burtenshaw and Sergio Paniego and Vaibhav Srivastav and Thibaud Frere and Ed Beeching and Lewis Tunstall and Leandro von Werra and Thomas Wolf},
  year={2025},
  howpublished={\url{https://huggingface.co/spaces/HuggingFaceH4/on-policy-distillation}},
}

@misc{niu2026breakingtokenizerbarrieronpolicy,
      title={Breaking the Tokenizer Barrier: On-Policy Distillation across Model Families}, 
      author={Yifan Niu and Han Xiao and Dongyi Liu and Zelong Wang and Dihong Gong and Yasheng Wang and Jia Li},
      year={2026},
      eprint={2606.09456},
      archivePrefix={arXiv},
      primaryClass={cs.LG},
}

@article{sun2026simctrecoveringlostsupervision,
      title={SimCT: Recovering Lost Supervision for Cross-Tokenizer On-Policy Distillation}, 
      author={Jie Sun and Mao Zheng and Mingyang Song and Qiyong Zhong and Yilin Cheng and Bichuan Feng and Pengfei Liu and Junfeng Fang and Xiang Wang},
      year={2026},
      eprint={2605.07711},
      archivePrefix={arXiv},
      primaryClass={cs.CL}
}
